\pdfoutput=1
\documentclass{article}
\usepackage[utf8]{inputenc}
\usepackage{authblk}
\usepackage[numbers]{natbib} 
\usepackage{xcolor}
\usepackage{tabularx}
\usepackage{amsmath,amssymb,amsthm}
\usepackage{graphicx}
\usepackage{fullpage}
\usepackage{enumitem}
\usepackage{multirow}
\usepackage{multicol}
\usepackage{mdframed}
\usepackage{makecell}
\usepackage{subcaption}
\usepackage[flushleft]{threeparttable}
\usepackage{longtable}
\usepackage{tikz}
\usetikzlibrary{calc}
\usepackage{caption}  
\usepackage[margin=1in]{geometry}
\usepackage{booktabs}
\usepackage{siunitx} 
\usepackage{url}

\title{
Size is Not the Solution: Deformable Convolutions for \textit{Effective} Physics Aware Deep Learning}
\author[1]{Jack T. Beerman}
\author[2]{Shobhan Roy}
\author[2]{H.S. Udaykumar}
\author[1,3]{Stephen S. Baek}

\affil[1]{School of Data Science, University of Virginia, Charlottesville, VA 22903, United States}
\affil[2]{Department of Mechanical Engineering, University of Iowa, Iowa City, IA 52242, United States}
\affil[3]{Department of Mechanical and Aerospace Engineering, University of Virginia, Charlottesville, VA 22903, United States}

\begin{document}

\maketitle

\begin{abstract}
Physics-aware deep learning (PADL) enables rapid prediction of complex physical systems, yet current convolutional neural network (CNN) architectures struggle with highly nonlinear flows. While scaling model size addresses complexity in broader AI, this approach yields diminishing returns for physics modeling. Drawing inspiration from Hybrid Lagrangian-Eulerian (HLE) numerical methods, we introduce deformable physics-aware recurrent convolutions (D-PARC) to overcome the rigidity of CNNs. Across Burgers' equation, Navier-Stokes, and reactive flows, D-PARC achieves superior fidelity compared to substantially larger architectures. Analysis reveals that kernels display anti-clustering behavior, evolving into a learned ``active filtration" strategy distinct from traditional h- or p-adaptivity. Effective receptive field analysis confirms that D-PARC autonomously concentrates resources in high-strain regions while coarsening focus elsewhere, mirroring adaptive refinement in computational mechanics. This demonstrates that physically intuitive architectural design can outperform parameter scaling, establishing that strategic learning in lean networks offers a more effective path forward for PADL than indiscriminate network expansion.
\end{abstract}

\section{Main}
In artificial intelligence, performance gains have largely followed scaling laws~\cite{DBLP:journals/corr/abs-2001-08361, DBLP:journals/corr/abs-1905-11946, Sevilla_2022}. While effective for vision and language tasks, this approach yields limited returns in physics-aware deep learning (PADL). In the physical systems of interest, complexity arises not from semantic diversity but from strongly nonlinear, dynamic and localized multiscale phenomena. As we show, deeper model counterparts fail to resolve sharp gradients and evolving interfaces despite consuming orders of magnitude more resources. Rather than brute-forcing performance through scale, we seek a more intelligent learning strategy by drawing inspiration from classical numerical methods in computational mechanics, which we demonstrate in this work. 

The backbone of most PADL models is a CNN, which is the natural choice to learn from field data or ``moving images.” However, ``pixel-based” CNNs originate from the computer vision community and require that physical quantities be resolved on a fixed Cartesian grid–mirroring the fixed-grid discretization of Eulerian solvers in computational mechanics. Eulerian approaches are not effective in capturing strongly localized nonlinear phenomena (such as shocks, interfaces, or boundary layers) because their uniform sampling cannot adapt to evolving sharp gradients or complex moving interfaces. While increasing network depth allows the model to aggregate information from a wider spatial context (i.e., a larger receptive field), it does not enable the adaptive spatial refinement necessary to resolve critical, localized, evolving features.

As an alternative, drawing from similar solutions from the computational physics field, a particle-based, or \textit{Lagrangian}, approach might be more desirable for predicting such strongly nonlinear phenomena. Typically graph neural networks (GNNs) may be employed as a Lagrangian equivalent of neural networks. Such a formulation naturally tracks the history of material elements, making it particularly effective for advection-dominated problems where maintaining historical context is crucial. However, particle-based formulations operate on unstructured point clouds without fixed connectivity requiring dynamic neighbor searches to determine spatial proximity~\cite{sanchez2020learning}. They can also be plagued by loss of resolution or excessive clustering of points that need to be periodically readjusted to maintain solution quality, making them significantly harder to utilize than grid-based Eulerian methods. This motivates hybridizing Eulerian and Lagrangian principles, combining their strengths.

Early numerical methods faced similar trade-offs between fidelity in Eulerian schemes ~\cite{laney1998computational, SHU1988439} and mesh distortion challenges in Lagrangian approaches~\cite{johnson2009numerical}, which motivated the development of Hybrid Lagrangian-Eulerian (HLE) frameworks such as Particle-in-Cell (PIC) and Arbitrary Lagrangian Eulerian (ALE) methods and immersed boundary methods (IBM). These approaches demonstrated robust performance and efficiency across large deformation problems~\cite{osti_4769185, brackbill1988flip, sulsky1994particle, bardenhagen2004generalized, Eiris2023} by retaining a fixed Eulerian grid as a computational ``scratch pad,” while particles transport material state and adaptively sample evolving features.

Here, we seek an HLE-inspired formulation of PADL, which combines the computational efficiency of Eulerian, pixel-based CNN formulations with the adaptivity of Lagrangian particle tracking in negotiating localized nonlinear phenomena. We hypothesize this yields a more accurate, efficient, and explainable PADL model relative to fixed Eulerian models of much larger size.

\begin{figure}
    \centering
    \includegraphics[width=\linewidth]{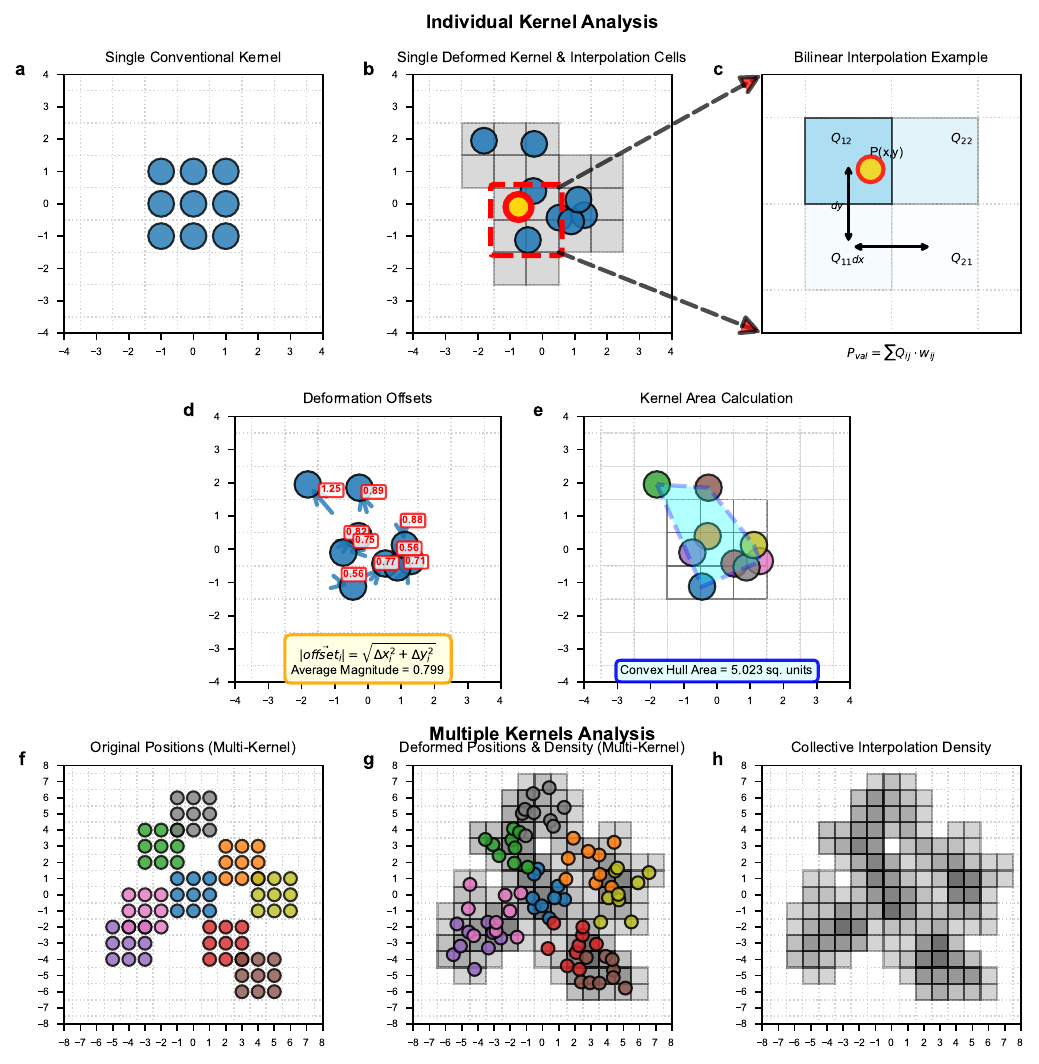}
    \caption{\textbf{Deformable convolution mechanics and collective scaffolding in D-PARC. Individual Kernel Analysis (a-e):} \textbf{a)} Fixed $3 \times 3$ convolution kernel with uniform sampling at grid points. \textbf{b)} Deformable kernel with learned spatial offsets (blue circles) and interpolation cells (gray squares); gold element highlighted for detailed analysis. \textbf{c)} Bilinear interpolation for the highlighted element, showing weighted contributions from four neighboring grid cells ($Q_{11}$, $Q_{21}$, $Q_{12}$, $Q_{22}$). \textbf{d)} Offset vectors (blue arrows) quantifying displacement from fixed to deformed positions; formula shows magnitude calculation. \textbf{e)} Convex hull area (blue shading) measuring spatial extent of the deformed kernel. \textbf{Multiple Kernels Analysis (f-h):} \textbf{f)} Original positions of nine independent kernels, each color-coded for tracking. \textbf{g)} Deformed positions of all kernels overlaid on interpolation density heatmap, showing spatial redistribution. \textbf{h)} Collective interpolation density map quantifying the cumulative sampling intensity across all kernels; darker regions indicate computational resource concentration where multiple deformed kernels cluster.}
    \label{fig:kernel_analysis}
\end{figure}

In this vein, we posit that \textit{\textbf{deformable convolutions}}~\cite{dai2017deformable} enable HLE behavior within PADL. Unlike fixed-kernel convolutions, deformable convolutions allow each element of a convolution kernel to move independently by learning \textit{spatial offsets}, enabling kernels to dynamically reshape their receptive fields and ``cluster” computational resources around salient features (\textbf{Figure~\ref{fig:kernel_analysis}}).

We introduce \textit{deformable convolutions} in \textit{physics-aware recurrent convolutions} (D-PARC)~\cite{nguyen2024parcv2, cheng2024physics} and demonstrate superior performance over larger fixed-kernel models. By establishing that physically intuitive architectural design outperforms indiscriminate network scaling, we provide a path forward for PADL that prioritizes adaptive computational allocation over brute-force parameterization.

Initially, we hypothesized that deformable kernels would function analogously to HLE methods, where the Lagrangian particles physically transport material states across the Eulerian grid, thereby exhibiting clustering in regions of heightened dynamics. However, surprisingly, our kinematic analysis of the kernel offsets reveals behavior contrary to this intuition drawn from experience with HLE methods. While the kernels interact with evolving fields, they do not emulate the strict Lagrangian transport as anticipated; instead, their motion diverges from classical particle tracking.

This prompted investigating the effective receptive field (ERF) of the neural network where we discover a learned strategy of ``active filtration.”  D-PARC adapts learning through selective physical context: concentrating intense focus by recruiting multiple neighboring convolutional kernels to scaffold representation of complex local features in high strain regions while effectively ``coarsening” its computations in low-strain areas. Thus, PADL models with deformable convolutions appear to succeed by \textit{seeing more with less}, effectively harnessing a notion of spatial selectivity emerging from the iterative scaffolding of the convolution operation. Altogether, this work underscores that intelligently-designed network architectures grounded in physical principles are superior to the brute-force scaling of models in PADL. 

\section{Results}

\subsection*{Model Size is Not the Solution}

We first test the hypothesis that deformable convolutions enable PADL models to resolve high-fidelity features more accurately than simple scaling approaches. To do so, we establish the PARCv2 model, as our primary baseline, which has been shown to outperform other PADL models in capturing non-linear flow fields with limited data ~\cite{cheng2024physics}. Alongside this baseline, we introduce the deeper variant (PARCv2-L), to serve as our scaled model with a significantly larger (spatial) context window than PARCv2. In computer vision, this ``context window” is defined as the theoretical receptive field (TRF) which quantifies the maximum spatial context the network can utilize. While PARCv2 is limited by its original implementation, PARCv2-L’s TRF spans the entire input field.

D-PARC maintains the same architecture size as PARCv2 but augments it with deformable convolutions enabling adaptive receptive fields through learned spatial offsets. We compare prediction accuracy, parameter count, and ERF sparsity across PARCv2, PARCv2-L, and D-PARC to isolate whether HLE-inspired design outperforms scaling.

We benchmark across three datasets of continuum flow calculations with diverse types of physics: (a) Inviscid compressible flow governed by the canonical Burgers’ equation (Supplementary~\ref{app:appendix_burgers}), (b) Incompressible Navier-Stokes flow at Re 20 to 1000 over a bluff body (Supplementary~\ref{app:appendix_ns}), and (c) Shock- induced collapse of pores in microstructures and consequent hotspot growth in a solid energetic material (Supplementary~\ref{app:appendix_em}). This selection is strategic, intended to evaluate whether deformable convolutions offer benefits that scale with physical complexity in a manner consistent with HLE methods.

For Burgers' equation (Table~\ref{tab:compact_dataset_metrics}), D-PARC shows marginal improvement (0.36\% RMSE reduction) versus PARCv2-L (1.8\%), insufficient to justify 1015.94\% more parameters.

For Navier-Stokes, D-PARC reduces velocity RMSE by 9.61\% and pressure RMSE by 1.56\%, significantly outperforming PARCv2-L (2.32\%, 0.42\%). As high-gradient transients (shocks, interfaces, eddies) dominate, adaptive sampling outperforms context window expansion.

This observation becomes more pronounced for EM, where the physical regime shifts significantly. Here, shock-pore interactions generate intense gradients along reaction fronts~\cite{nguyen2022multi}. D-PARC reduces velocity magnitude RMSE by 5.2\%, pressure RMSE by 4.8\%, and temperature RMSE by 0.8\%. In contrast, PARCv2-L degrades performance, increasing velocity magnitude RMSE by 3.5\% and temperature RMSE by 57\%.

\begin{table}[ht!]
\centering
\small 
\renewcommand{\arraystretch}{1.2}
\setlength{\tabcolsep}{2pt} 

\begin{tabularx}{\textwidth}{l @{\hspace{1cm}} >{\centering\arraybackslash}X >{\centering\arraybackslash}X >{\centering\arraybackslash}X c}
\hline
\textbf{Model} & \textbf{RMSE (U)} & \textbf{RMSE (P)} & \textbf{RMSE (T)} & \textbf{Params (M)} \\
\hline
\multicolumn{5}{l}{\textbf{BURGERS}} \\
\hline
PARCv2   & 0.0189  & --- & --- & 1.38 \\
PARCv2-L & 0.0185 (\textcolor{blue}{+1.8\%}) & --- & --- & 15.4 \\
D-PARC   & 0.0188 (\textcolor{blue}{+0.36\%}) & --- & --- & 1.49 \\
\hline
\multicolumn{5}{l}{\textbf{NAVIER–STOKES}} \\
\hline
PARCv2   & 0.266 & 8.61$\times$10$^{2}$  & --- & 16.3 \\
PARCv2-L & 0.259 (\textcolor{blue}{+2.3\%}) & 8.57$\times$10$^{2}$ (\textcolor{blue}{+0.4\%}) & --- & 240.6 \\
D-PARC   & 0.24 (\textcolor{blue}{+9.6\%}) & 8.47$\times$10$^{2}$ (\textcolor{blue}{+1.5\%}) & --- & 16.8 \\
\hline
\multicolumn{5}{l}{\textbf{EM(s)}} \\
\hline
PARCv2   & 2.05$\times$10$^{2}$  & 1.93$\times$10$^{9}$  & 285 & 19.2 \\
PARCv2-L & 2.12$\times$10$^{2}$ (\textcolor{red}{-3.5\%}) & 1.84$\times$10$^{9}$  (\textcolor{blue}{+4.5\%}) & 448 (\textcolor{red}{-57\%}) & 243.5 \\
D-PARC   & 1.95$\times$10$^{2}$ (\textcolor{blue}{+5.2\%}) & 1.83$\times$10$^{9}$  (\textcolor{blue}{+4.8\%}) & 283 (\textcolor{blue}{+0.8\%}) & 19.7 \\
\hline
\end{tabularx}
\vspace{4mm}
\noindent

\begin{tabularx}{\textwidth}{l @{\hspace{1cm}} >{\centering\arraybackslash}X >{\centering\arraybackslash}X >{\centering\arraybackslash}X >{\centering\arraybackslash}X c} 
\multicolumn{6}{l}{\textbf{EM (Hotspot Metrics)}} \\
\hline
\textbf{Model} & \textbf{IoU} & \textbf{Dice} & \textbf{RMSE (K)} & \textbf{W-RMSE (K)} & \textbf{Params (M)} \\
\hline
PARCv2 & 0.341 & 0.486 & 1177 & 1907 & 19.2 \\
PARCv2-L & 0.230 (\textcolor{red}{-33\%}) & 0.357 (\textcolor{red}{-27\%}) & 1192 (\textcolor{red}{-1.2\%}) & 2102 (\textcolor{red}{-10\%}) & 243.5 \\
D-PARC & 0.382 (\textcolor{blue}{+12\%}) & 0.533 (\textcolor{blue}{+9.6\%}) & 1160 (\textcolor{blue}{+1.5\%}) & 1847 (\textcolor{blue}{+3.2\%}) & 19.7 \\
\hline
\end{tabularx}

\caption{\textbf{D-PARC outperforms fixed-kernel models across flow regimes of increasing complexity.} Prediction accuracy measured by root mean square error (RMSE) for velocity magnitude ($U$), temperature ($T$), and pressure ($P$) across three datasets: Burgers' equation (inviscid compressible flow), Navier-Stokes (incompressible flow over bluff body, Re 20-1000), and energetic material (EM) with shock induced pore collapse. PARCv2 serves as baseline; PARCv2-L is the deeper scaled variant with 10x more parameters; D-PARC incorporates deformable convolutions at comparable size to PARCv2. Percentages in parentheses indicate relative improvement (blue, positive) or degradation (red, negative) versus PARCv2. For EM dataset, additional metrics quantify hotspot geometry ($T > 875$ K): Intersection over Union (IoU) measures spatial overlap between predicted and ground truth hotspot regions (0 = no overlap, 1 = perfect), Dice score, RMSE measuring temperature error within correctly identified hotspot pixels; W-RMSE (K) is IoU weighted temperature error penalizing incorrect hotspot locations. D-PARC achieves superior accuracy in complex regimes (Navier-Stokes, EM) while PARCv2-L shows diminishing or negative returns despite 12.6x parameter increase. Parameter counts (Params) listed in millions (M).}
\label{tab:compact_dataset_metrics}
\end{table}

The performance difference becomes even more dramatic when examining hotspot geometry—a critical quantity for predicting material ignition~\cite{welle2014microstructural}. RMSE can be minimized by smoothing. Therefore, we prioritize IoU and Dice scores (Supplementary~\ref{app:hotspot_metrics}) which penalize failures to resolve evolving hotspots. We focus on EM for remaining results given its high nonlinearity, while noting similar behavior is exhibited in all datasets and further analyzed in Supplementary~\ref{app:appendix_burgers}and~\ref{app:appendix_ns}.

D-PARC achieves a relative improvement of $+12.0\%$ in IoU and $+9.6\%$ in Dice score over PARCv2 in hotspot reconstruction. In contrast, PARCv2-L suffers in performance (IoU $-32.7\%$, Dice $-26.7\%$). This demonstrates depth alone fails for multi-scaled coupled phenomena. We hypothesize that in the context of nonlinear physical dynamics, depth may exacerbate optimization without addressing Eulerian rigidity. To go deeper, we investigate this relationship between deformable convolutions and HLE mechanics in the following section and unearth interesting and counterintuitive insights.

\subsection*{The Learned Lagrangian: How Deformable Convolutions Capture Flow Fields}

While quantitative results confirm that D-PARC outperforms fixed-kernel models, the question remains: \textit{how do deformable convolutions achieve this?} Our architectural motivation for D-PARC was predicated on the hypothesis that deformable convolutions will naturally emulate HLE methods, with kernel elements following physical material advection through the domain and clustering in regions of high gradients, resulting in improved resolution of such regions. To test the validity of our design philosophy, we now visualize and quantify how kernel elements dynamically reallocate in response to high- and low-strain regimes and assess the degree to which kernel behavior aligns with the adaptive sampling strategies of traditional HLE solvers.

To illustrate the learned behavior, we visualized the kernel trajectories at high- and low-strain regimes. The strain rate quantifies the local deformation, serving as a diagnostic for spatial gradients in the velocity field, $u(\mathbf{r}, t)$, and for identifying regions where kernel transport behavior is influenced by local deformation dynamics.
 
\begin{figure}
    \centering
    \includegraphics[width=\linewidth]{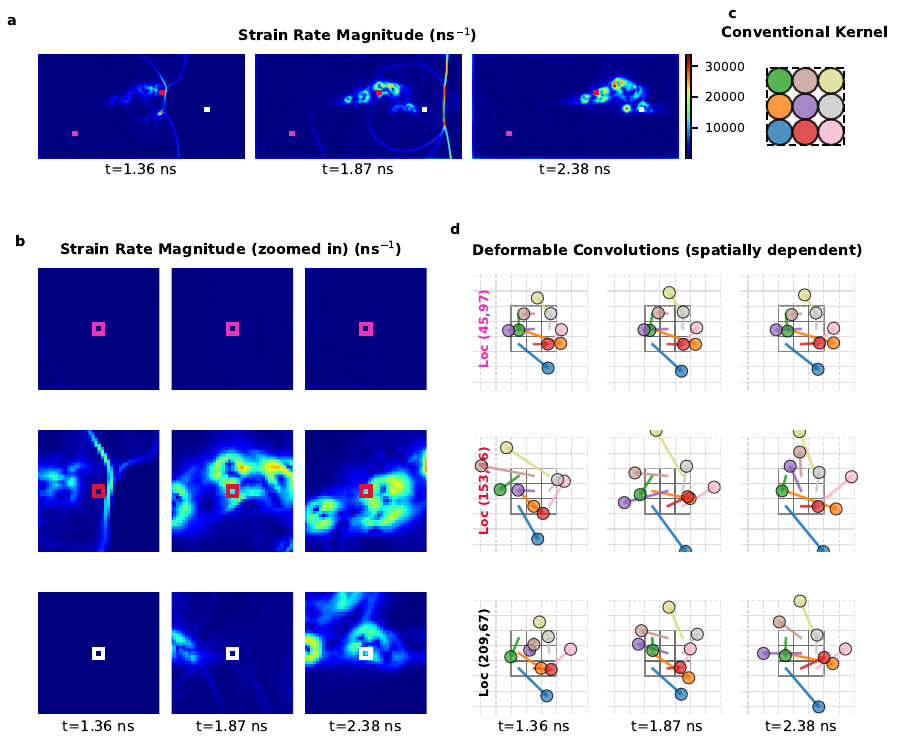}
    \caption{\textbf{Deformable kernels dynamically adapt receptive fields in response to local strain dynamics.} Visualization of D-PARC kernel behavior during shock-induced pore collapse in EM simulation \textbf{a)} Strain rate magnitude field ($\dot{\epsilon}$) at three timesteps (t = 1.36ns, 1.87ns, 2.38ns) following shock passage. Colored squares mark three spatial locations: Loc(45,97) in low strain region (magenta), Loc(153, 46) at shock front (red), and Loc (209, 67) ambient to high strain (white). \textbf{b)} Zoomed views ($40\times40$ pixel regions) of the three locations, revealing localized strain patterns; colored outline indicate $3\times3$ kernel with nine elements. \textbf{c)} Conventional $3\times3$ convolution kernel with nine elements (colored circles) at fixed grid positions. \textbf{d)} Deformed kernel configurations at the three locations across timesteps. Colored circles represent kernel element positions after learned spatial offsets; arrows show displacement vectors from fixed positions (panel c). Gray gridlines indicate underlying Eulerian grids. Kernels expand substantially in high-strain regions (e.g., Loc (153, 46) at t=1.87ns) while remaining compact in low-strain regions (Loc (45, 97), demonstrating adaptive context window based on local flow dynamics.}
    \label{fig:spatial_deformable_kernels}
\end{figure}

\textbf{Figure~\ref{fig:spatial_deformable_kernels}} illustrates the EM dataset at three locations for \(t = 1.36-2.38 ns\), immediately following pore collapse. Each kernel is anchored at its center on the Eulerian grid, and this \textit{anchor point} serves as the reference from which kernel elements are transported across the domain. From Figure~\ref{fig:spatial_deformable_kernels}, D-PARC dynamically expands its context window for individual kernels in volatile regions, such as shock and burn fronts, while conserving resources in smoother regions. For example, the deformed kernel at Loc(153,46) develops a receptive field of a substantial spread at \(t = 1.87 ns\) when the anchor point encounters a shockwave. Visualization of representative kernels for the other datasets are presented in Supplementary~\ref{app:ind_an}.

We recorded for each kernel, the strain rate magnitude at its anchor point, the displacement magnitude of its nine elements relative to fixed-kernel positions, and the total area spanned. In regions of high strain, kernel elements displaced  \textit{farther from their anchors and encompassed larger areas}; in low-strain regions displacements remained minimal (Supplementary Figure~\ref{fig:strain_statistics}). Supplementary~\ref{app:ind_an} shows the plotted metrics for all datasets, demonstrating D-PARC distinguishes between high and low strain regimes, as majority of the kernels expand their receptive field and traverse far from the anchor point in regions of higher strain. This is \textit{opposite} of HLE methods, where the Lagrangian particles cluster in regions of steep gradients such as shock fronts, boundary layers and isolated eddies, while becoming sparse in extensional regions~\cite{song2021improved}. Below, we shed light on these seemingly counterintuitive results by analyzing the ``collective scaffolding'' of deformable convolutions. This mechanism describes the dense clustering of kernel offsets, effectively stacking multiple sampling points onto physically critical features. 

\begin{figure}
    \centering
    \includegraphics[width=\linewidth]{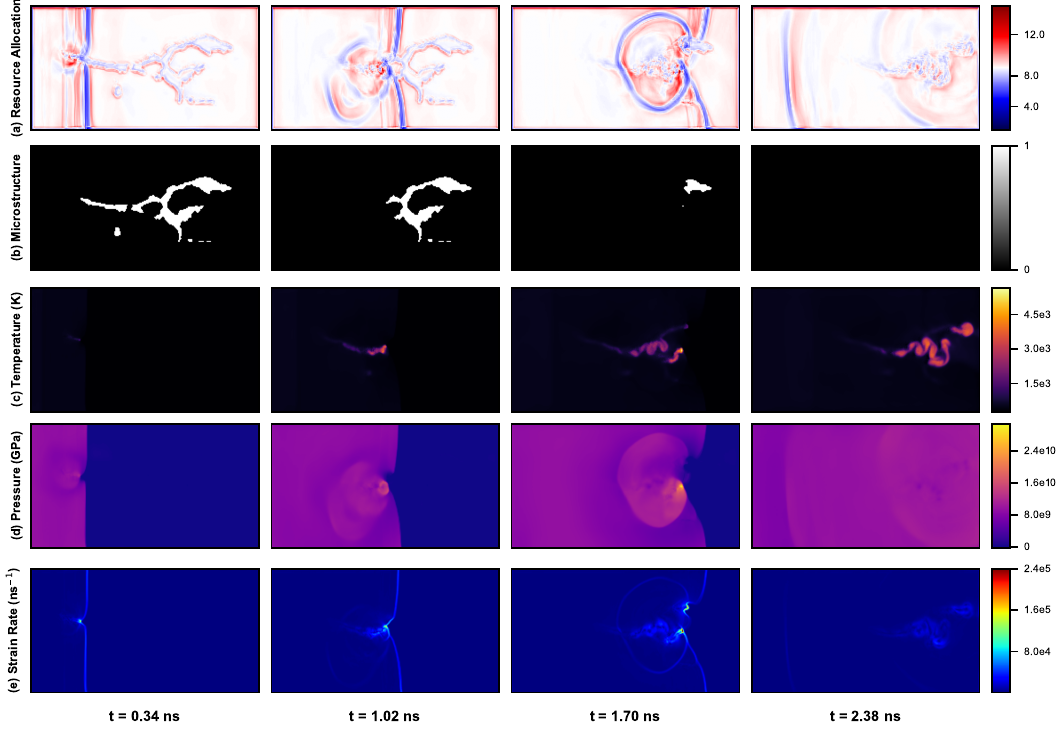}
    \caption{\textbf{D-PARC concentrates computational resources on physically salient flow features.} Temporal evolution of resource allocation and physical fields during shock-induced pore collapse in energetic material simulation across four timesteps ($t = 0.34$, $1.02$, $1.70$, $2.38$~ns). \textbf{a}, Learned resource allocation quantified by cumulative interpolation density (number of times each pixel is sampled by deformed kernels). Diverging colormap centered at 9 (baseline for conventional $3\times3$ convolution sampling each pixel nine times); values $>9$ (red) indicate computational hotspots where D-PARC clusters multiple kernels. \textbf{b}, Binary microstructure map (white = solid material, black = void/pore). \textbf{c}, Temperature field (K) showing hotspot formation and growth. \textbf{d}, Pressure field (GPa) displaying shock propagation and material response. \textbf{e}, Strain rate magnitude (ns$^{-1}$) quantifying local deformation rates. D-PARC autonomously concentrates resources (panel~a) on pore boundaries, incoming shock fronts, and emerging blast waves---regions exhibiting steep gradients in temperature, pressure, and strain rate. Resource allocation intensifies at leading edges where gradients are steepest while remaining minimal in smooth regions and domain boundaries. No prior physical guidance was provided; allocation patterns emerge purely from data-driven learning.}
\label{fig:resource_allocation}
    \label{fig:feature_evolution}
\end{figure}

To understand where D-PARC allocates its resources, we visualize spatial clustering \textbf{Figure \ref{fig:feature_evolution}}. Deformable convolutions use bilinear interpolation between grid points, meaning some pixels contribute more to predictions than others. We quantify this by counting how often each pixel is sampled: summing the interpolation weights at each Eulerian location.

A standard $3 \times 3$ convolution samples each pixel exactly nine times. We use this as our baseline. The color scale in \textbf{Figure \ref{fig:feature_evolution}} is normalized to nine. Regions higher than this baseline indicate regions where deformable kernels have clustered together, revealing where D-PARC prioritizes its representational capacity to resolve challenging physics.

D-PARC concentrated resources on the boundaries of the pore (EM-void interface), incoming shock front, and blast wave. This aligns with the physics of the problem which highlights the role of energy localization due to interaction of microstructural heterogeneities with overpassing shock in the formation of hotspots that act as precursors to shock-to-detonation transition (SDT)~\cite{welle2014microstructural}. Across all datasets, D-PARC similarly devotes resources to regions of salient flow features: shock and reaction fronts, shear layers and vortices. 

Resource allocation was non-uniform. Computation intensified at the leading edge of shocks or fronts where gradients are steepest while effort decreased in smooth regions behind the shock front. Minimal resources were allocated to domain boundaries which contribute little to global dynamics. This resembles the behavior of particle clustering and was only achieved via deformable convolutions. We note that the scaffolding behavior with high cumulative kernel coverage is different from the working of individual kernels where the context window widens with spread-out kernel elements in similar high-gradient regions.

Hence the seemingly counterintuitive results above, in fact, rise from the fact that the analogy between HLE and D-PARC is one of \textit{conceptual resemblance} rather than methodological equivalence. Both frameworks embody the principle of adaptive sampling over a fixed, structured background field. In particle-in-cell (PIC), the grid-to-particle (G2P) and particle-to-grid (P2G) interpolation steps transfer quantities between representations. In D-PARC, these ideas are loosely mirrored: kernel elements move in a Lagrangian manner across the domain and are updated by bilinear interpolation from the underlying structured DNS data, then network weights
are applied and results are projected back onto a fixed  \((H \times W)\) Eulerian field. The differences stem from the fundamental contrast between numerical solvers and neural networks: in HLE, particle motion occurs under the influence of explicit physical laws (e.g., conservation of momentum and energy) whereas in D-PARC, kernel movement is optimized through backpropagation. Thus, while both frameworks relax the rigidity of fixed spatial sampling to better capture localized dynamics, HLE achieves this through physics-based laws, whereas D-PARC achieves this through data-driven learning.

Figures~\ref{fig:spatial_deformable_kernels} and \ref{fig:feature_evolution} reveal key observations. First, HLE-type methods tend to cluster in shear zones and eddies under the influence of the governing flow equations rather than an explicit allocation strategy. This clustering is therefore incidental and often detrimental: it can lead to local over- or under-sampling and trigger numerical instabilities.. In contrast, the effective clustering observed in D-PARC is a learned mechanism for redistributing representational resources toward dynamically important regions, rather than an uncontrolled consequence of the dynamics. Second, flow computations where such clustering does occur at discontinuities, high-order schemes for flux reconstruction are typically downgraded or supplemented with artificial viscosity to preserve monotonicity and prevent spurious oscillations~\cite{toro2013riemann}. Individual kernels in D-PARC, tend to expand their receptive fields in regions of high gradients to enhance overall feature extraction, often yielding overlaid kernel point trajectories and skewed kernel shapes (Figure~\ref{fig:spatial_deformable_kernels}). 

Unlike in computational mechanics, D-PARC does not exhibit numerical instabilities associated with anisotropic clustering because their adaptive sampling is governed by optimization objectives rather than by conservation and entropy constraints of an underlying physical PDE. Rather, the physical constraints are inherited by the model when being trained on DNS data of the \textit{solved} flow field. These distinctions underscore that, while DCNs embody HLE principles, their role is to learn to  \textit{emulate} rather than \textit{calculate} physical dynamics.

Thus, these observations contrast our initial design philosophy for D-PARC and purpose for incorporating deformable convolutions. Cumulatively, they suggest that the model’s success is not merely a result of tracking particles but rather stems from a more intricate advantage in how it gathers contextual information. To understand this, we must look beyond the kinematic analysis of deformable convolutions and analyze the information flow itself. This reveals D-PARC’s advantage lies not in tracking particles, but in learned resource allocation – examined next through ERF analysis.

\subsection*{Seeing More with Less: \textit{Effective} Physics Aware Deep Learning}\label{sec:erf}

To determine the source of the advantage of deformable convolutions in PADL, we distinguish between their capacity to gather or track information and the actual  \textit{influence} of the information gathered.In traditional CNNs, increasing model depth for a larger context window requires indiscriminately processing all pixels. D-PARC, appears to decouple context from spatial coverage. We quantify this efficiency through the ERF, which measures the actual gradient of influence of each input pixel (i.e., physical field variable) on the predicted dynamics ~\cite{luo2016understanding}.

While kernel transport depicts where the model is allocating its computational budget, this only represents the \textit{potential} for information transfer. The ERF allows us to measure the physical intuition, identifying what context such as flow features or shock fronts, the model deems essential for predicting spatiotemporal dynamics. We can infer D-PARC’s selection strategy by analyzing the  \textbf{sparsity} of the ERF. In PARCv2, the ERF typically manifests as a diffuse Gaussian distribution, indicating that the model relies on a broad, indiscriminate aggregation of local information~\cite{luo2016understanding}. In contrast, D-PARC reveals a sparser ERF structure where gradients are not spread uniformly. Instead, the ERF adapts to distinctly associate high- and low-strain regions. If D-PARC achieves superior accuracy while effectively  ``ignoring” the vast majority of its context window relative to the fixed models, we can deduce that the model has learned to filter out physical redundancy and unnecessary context. 

Following gradient-based receptive field analysis~\cite{luo2016understanding}, we compute the \textbf{effective area} as the number of spatial locations whose maximum gradient magnitude exceeds threshold $\tau$ (we test $\tau \in [0.5\%, 10\%]$ of maximum gradient). Defining sparsity relative to domain size \(H \times W\) is misleading when comparing models with different depths. A shallow model with a limited field of view (TRF) would
appear artificially ``sparse" simply because it is incapable of observing the broader domain thus conflating architectural blindness with spatial selectivity. 

To address this, we define \textit{spatial sparsity} as the percentage of the model’s available capacity that is effectively utilized:

\begin{equation}
\text{Spatial Sparsity}(\tau) = \left(1 - \frac{\text{Effective Area}(\tau)}{\min(\text{TRF}, H \times W)}\right) \times 100\%
\end{equation}

Here, \(min(TRF, H \times W)\), represents the maximum possible spatial context the model is theoretically capable of accessing within the domain boundaries. 

A higher sparsity score indicates that the model possesses the capacity to observe a broad context but actively filters (\textbf{\textit{active filtration}}) its computation on a restricted subset of relevant features. In Table~\ref{tab:erf_threshold_sensitivity}, we report these metrics at multiple gradient thresholds $\tau$. We analyze representative snapshots at $T=6$ that drive the prediction of the state at $T=7$. We perform this analysis across six distinct spatial locations (three in high-strain regions where rapid evolution occurs, and three in low-strain regions with slower dynamics).

D-PARC demonstrates superior selectivity in both high- and low-strain regions, achieving the highest sparsity of all models while utilizing substantially less information than PARCv2-L in both flow regimes (Table ~\ref{tab:erf_threshold_sensitivity}). This reflects ``active filtration”: deformable convolutions learn where to look, while network weights refine what to extract.

\begin{table}[!ht]
\centering
\small
\begin{tabular}{lcccccc}
\toprule
\multirow{2}{*}{\textbf{Thresh.}} & \multicolumn{2}{c}{\textbf{D-PARC}} & \multicolumn{2}{c}{\textbf{PARCv2}} & \multicolumn{2}{c}{\textbf{PARCv2-L}} \\
\cmidrule(lr){2-3} \cmidrule(lr){4-5} \cmidrule(lr){6-7}
\textbf{(\%)} & \textbf{Area} & \textbf{Spar. (\%)} & \textbf{Area} & \textbf{Spar. (\%)} & \textbf{Area} & \textbf{Spar. (\%)} \\
\midrule
\multicolumn{7}{l}{\textit{Theoretical RF: PARCv2 = 6,084 px, D-PARC/PARCv2-L = 101,124 px}} \\
\midrule
\multicolumn{7}{l}{\textbf{High-Strain Regions} (n=3, mean±std)} \\
\midrule
0.5  & \textbf{4376±1906}   & \textbf{86.7±5.8}  & 4702±175   & 85.7±0.5 & 6251±379  & 80.9±1.2 \\
1.0  & \textbf{1836±1146}   & \textbf{94.4±3.5}  & 2758±232   & 91.6±0.7 & 3689±162  & 88.7±0.5 \\
2.0  & \textbf{475±337}     & \textbf{98.6±1.0}  & 948±194    & 97.1±0.6 & 1676±63   & 94.9±0.2 \\
3.0  & \textbf{194±109}     & \textbf{99.4±0.3}  & 414±106    & 98.7±0.3 & 903±48    & 97.2±0.2 \\
5.0  & \textbf{74±16}       & \textbf{99.8±0.1}  & 154±13     & 99.5±0.0 & 437±20    & 98.7±0.1 \\
7.5  & \textbf{45±14}       & \textbf{99.9±0.0}  & 79±4       & 99.8±0.0 & 329±13    & 99.0±0.0 \\
10.0 & \textbf{32±10}       & \textbf{99.9±0.0}  & 53±8       & 99.8±0.0 & 267±8     & 99.2±0.0 \\
\midrule
\multicolumn{7}{l}{\textbf{Low-Strain Regions} (n=3, mean±std)} \\
\midrule
0.5  & \textbf{1154±600}    & \textbf{96.5±1.8}  & 4889±2292  & 85.1±7.0 & 8092±6201 & 75.3±18.9 \\
1.0  & \textbf{492±335}     & \textbf{98.5±1.0}  & 3576±2435  & 89.1±7.4 & 4442±3322 & 86.5±10.1 \\
2.0  & \textbf{196±132}     & \textbf{99.4±0.4}  & 2343±2142  & 92.9±6.5 & 2113±1381 & 93.6±4.2 \\
3.0  & \textbf{125±70}      & \textbf{99.6±0.2}  & 1753±1719  & 94.7±5.3 & 1355±741  & 95.9±2.3 \\
5.0  & \textbf{71±33}       & \textbf{99.8±0.1}  & 1041±1126  & 96.8±3.4 & 788±319   & 97.6±1.0 \\
7.5  & \textbf{46±19}       & \textbf{99.9±0.1}  & 572±650    & 98.3±2.0 & 482±141   & 98.5±0.4 \\
10.0 & \textbf{35±15}       & \textbf{99.9±0.1}  & 351±401    & 98.9±1.2 & 326±59    & 99.0±0.2 \\
\bottomrule
\end{tabular}
\caption{\textbf{D-PARC achieves superior spatial selectivity through learned active filtration.} Effective receptive field (ERF) analysis quantifying spatial sparsity across gradient thresholds ($\tau$) in high-strain and low-strain regions of EM simulation at $T=6$ (predicting state at $T=7$). Effective Area measures the number of pixels whose maximum gradient magnitude exceeds threshold $\tau$ (reported as percentage of maximum gradient: 0.5\%, 1\%, 2\%, 3\%, 5\%, 7.5\%, 10\%). Sparsity quantifies efficiency of resource allocation, where higher sparsity indicates the model possesses capacity to observe broad context but actively filters computation to a restricted subset of relevant features. Theoretical receptive field (TRF): PARCv2 = 6,084 pixels; D-PARC and PARCv2-L = 101,124 pixels (entire domain). Metrics averaged over three spatial locations per regime ($n=3$, mean $\pm$ std). \textbf{High-strain regions:} shock fronts, pore boundaries, and reaction zones with rapid temporal evolution. \textbf{Low-strain regions:} smooth flow with slow dynamics. D-PARC achieves highest sparsity across all thresholds in both regimes, utilizing substantially fewer pixels than PARCv2-L (e.g., at $\tau=1\%$: D-PARC uses 492$\pm$335 pixels in low-strain vs.\ PARCv2-L's 4442$\pm$3322 pixels---9$\times$ more efficient). Critically, D-PARC dramatically reduces effective area in low-strain regions (1836 to 492 pixels at $\tau=1\%$), while PARCv2-L counterintuitively expands context where dynamics are simplest. Bold values indicate best performance per threshold.}
\label{tab:erf_threshold_sensitivity}
\end{table}

Initially, this may appear counterintuitive from a traditional computational physics perspective, where one might expect that accurately resolving high-gradient transients is handled by increasing local resolution or sampling density, while smoother regions are allowed to remain comparatively coarse. In contrast, D-PARC has high sparsity in both regimes which reflects a fundamental advantage of deformable convolutions in PADL. Moreover, the model learns to allocate more informative pixels in high- strain areas while dramatically filtering context in low-strain regions. This behavior parallels a core paradigm in computational mechanics: adaptive resource allocation based on local dynamics.

\begin{figure}
    \centering
    \includegraphics[width=\linewidth]{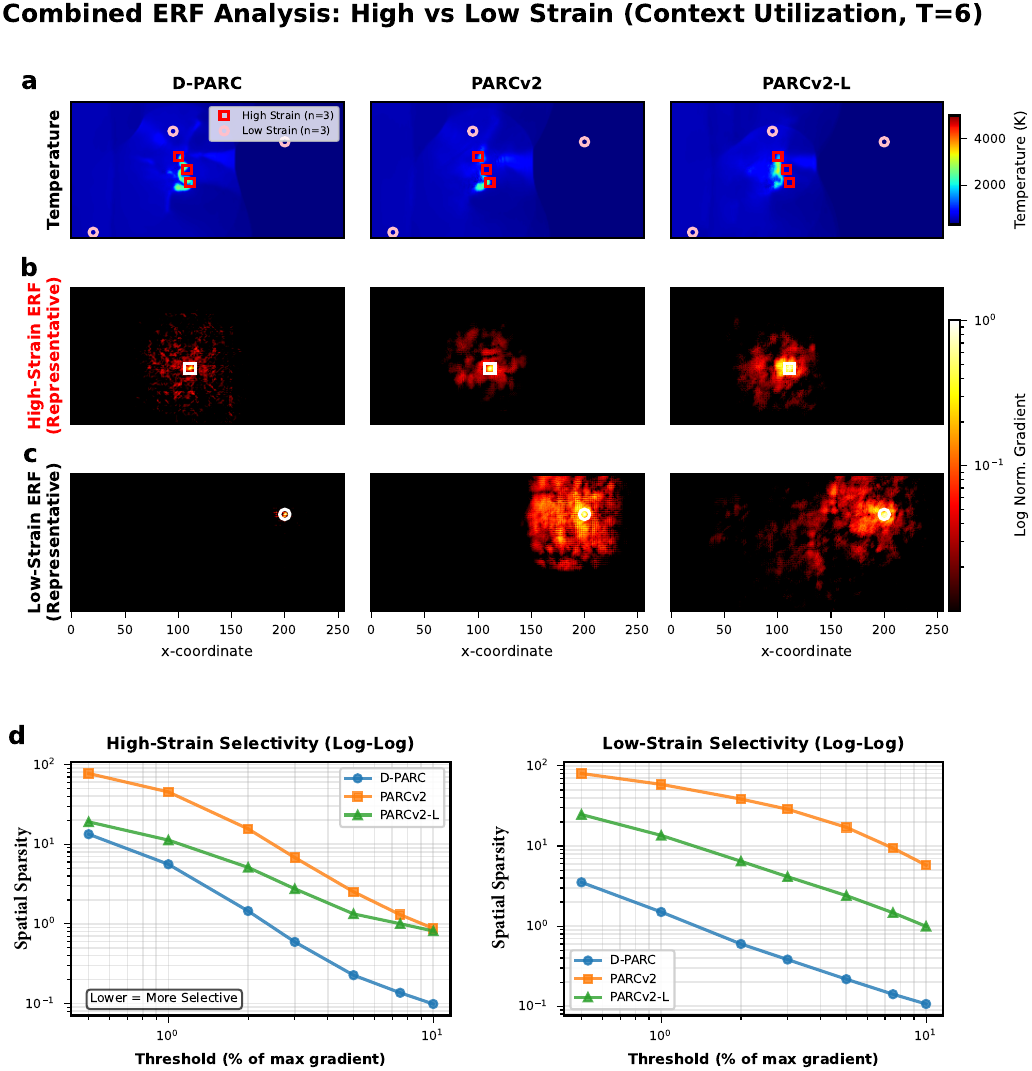}
    \caption{\textbf{D-PARC demonstrates regime-dependent \textit{active filtration} through spatially localized effective receptive fields.} Effective receptive field (ERF) visualization for energetic material simulation at $T=6$, comparing D-PARC, PARCv2, and PARCv2-L. \textbf{a}, Temperature field (K) with sampled locations: high-strain regions (red squares, $n=3$) and low-strain regions (white circles, $n=3$). \textbf{b}, Representative high-strain ERF heatmaps at gradient threshold $\tau=1\%$. Log-scale colormap shows normalized gradient magnitude; brighter regions indicate stronger influence. D-PARC exhibits compact ERF; PARCv2-L displays broad, diffuse influence. \textbf{c}, Representative low-strain ERF heatmaps at $\tau=1\%$. D-PARC contracts dramatically in smooth regions (492$\pm$335 pixels); PARCv2-L counterintuitively expands (4442$\pm$3322 pixels). \textbf{d}, Active filtration (log-log scale) across gradient thresholds (0.5--10\%). D-PARC achieves lowest sparsity (highest selectivity) in both regimes. Lower curves indicate more selective resource allocation. See Table~\ref{tab:erf_threshold_sensitivity} for complete metrics.}
    \label{fig:erf_viz}
\end{figure}

The most profound physical insight emerges in the low-strain regions. Physically, smooth flow regions possess lower information entropy; the dynamics are governed by simple dynamics or steady states, meaning one does not need massive ``context” to predict the next state. D-PARC captures this physical reality: its effective area drops drastically at \(\tau = 1\%\)  from 1,836 pixels in high-strain regions to just 492 pixels in low-strain regions (Figure~\ref{fig:erf_viz}).

PARCv2-L does the opposite. The model’s effective area expands in low-strain regions (lower sparsity), attempting to use more context and information where dynamics are simplest. At various gradient thresholds, PARCv2-L utilizes \textbf{9-10x} more pixels than D-PARC to resolve the same simple dynamics and paradoxically use more pixels (information) in low-strain regions than in high-strain regions. Unable to modulate local sampling density based on flow conditions, the fixed-architecture is forced to utilize pixels regardless of the spatiotemporal dynamics it exhibits. Consequently, they inefficiently aggregate uncorrelated background noise in smooth regions, resulting in lower sparsity and a misallocation of computational resources.

This analysis reconciles the divergence of deformable convolutions from HLE-based solvers observed in the previous section. While not tracking particles like HLE, D-PARC’s ERF demonstrates learned adherence to adaptive resource allocation widely incorporated in computational mechanics. By simultaneously discovering a strategy that concentrates influence on physical features while filtering unnecessary context, deformable convolutions achieve a physics-aware sparsity that fixed architectures cannot replicate. This validates that in PADL, performance depends not on the volume of context consumed, but on the precision of the context selection, demonstrating that selective information gathering is both computationally efficient and physically accurate.

\section{Discussion}

We challenge the notion that network size drives physics-aware deep learning performance. Across Burgers' equation, Navier-Stokes, and energetic materials, D-PARC's deformable convolutions—inspired by HLE methods, capture localized features more efficiently than deeper fixed-kernel networks. D-PARC does not strictly emulate HLE particle tracking but learns ``active filtration": concentrating resources in high-strain regions while coarsening them in low-strain areas, mirroring adaptive schemes without explicit refinement.

The path forward for PADL lies not in indiscriminate scaling but in architectures that encode domain-specific insight of the physical system being learned. D-PARC, demonstrates that networks can autonomously discover computational strategies analogous to classical numerical methods when given the ability to incorporate such physical principles. This shift from brute-force scaling to physics-informed strategic learning demonstrates that the most effective PADL models will continue to emerge from bridging computational physics intuition and theory with modern deep learning, rather than treating physical systems under the same umbrella heuristic approaches developed for vision and language tasks, thus opening new avenues for developing efficient, generalizable models for complex dynamical systems.

\section{Methods}

\subsection*{D-PARC Architecture}

To incorporate hybrid Lagrangian-Eulerian concepts in PADL, we model the advection-diffusion-reaction (ADR) equations using the PARC architecture, as previously accomplished by Nguyen et al.~\citep{nguyen2024parcv2}; however, we integrate \textit{deformable convolutions}~\cite{dai2017deformable} into PARC to emulate HLE-type behavior (Supplementary Figure~\ref{fig:DPARC_arch}). Unlike fixed convolutions that apply uniform spatial kernels, deformable convolutions learn continuous offset fields that dynamically adjust sampling locations during inference. This enables D-PARC to adaptively concentrate sampling density on salient features-shock fronts, reaction zones, phase boundaries while reducing resolution in smooth regions, analogous to how Lagrangian particles in PIC methods track moving structures rather than uniformly sampling space.

We consider a vector-valued function $\mathbf{u}(t,\mathbf{x})$, each of whose elements describes the time-evolution of a physical quantity $u^{(k)}$ as a mapping from $\mathcal{T} \oplus \Omega$ to $\mathbb{R}$. The superscript $k\in\{1,\cdots,N_c\}$ in parentheses indexes different physical quantities, such as temperature and pressure. The time $t$ is an element of the interval $\mathcal{T}:=[0, T]\subset\mathbb{R}^+$ and the position $\mathbf{x}$ is an element of the $d$-dimensional physical domain $\Omega \in \mathbb{R}^d$. The `$\oplus$' symbol represents the direct sum.

The evolution of the physical quantities $\mathbf{u}$ is assumed to be governed by some partial differential equations whose exact form is unknown:
\begin{equation}
    \partial_t \mathbf{u}(t,\mathbf{x}) = \mathcal{D}(\mathbf{u}; \lambda),
    \label{eq:generic_PDE}
\end{equation}
where $\mathcal{D}(\cdot,\lambda)$ is an unknown differential operator parametrized by physical parameters $\lambda$ (e.g., thermal diffusivity, kinematic viscosity). Our goal is to approximate the differential operator $\mathcal{D}$ using a convolutional neural network trained with a relevant data set.

For this work, we suppose that the training data is given on a fixed Euclidean (pixel-wise) grid, in which the function $\mathbf{u}(t, \mathbf{x}_i)$ is discretized at each pixel $\mathbf{x}_i$ for a time slice $t$. From the structure of the grid, each pixel $\mathbf{x}_i$ is surrounded by pixels $\mathbf{x}_j$ in its neighborhood $j \in \mathcal{N}_i$.

In this setting, we aim to learn the differential operator $\mathcal{D}$ in Equation~\ref{eq:generic_PDE} using convolutional neural networks. A convolution kernel $\mathcal{K}$ is considered to be an operator acting locally on the function $\mathbf{u}$, modeling spatial derivatives of $\mathbf{u}$.
Based on the universal approximation theorem for operators~\cite{chen1995universal}, the composition of such kernels with a sufficient depth and breadth should allow us to approximate the unknown differential operator $\mathcal{D}$ from data.

Specifically, we define the spatial domain $\Omega = [0, L_x] \times [0, L_y] \subset \mathbb{R}^2$, which is discretized into a uniform Cartesian grid $\mathcal{G} = \{ \mathbf{x}_{i,j} = (x_i, y_j) \mid i = 0, \dots, N_x; \, j = 0, \dots, N_y \}$. Here, $L_x$ and $L_y$ denote the physical extents of the domain in the $x$- and $y$-directions, respectively, and $N_x$, $N_y$ denote the number of grid intervals along each axis. The corresponding grid spacings are given by $\Delta x = L_x / N_x$ and $\Delta y = L_y / N_y$, and each pixel $\mathbf{x}_{i,j}$ corresponds to a fixed spatial location in a 2D Euclidean frame.

Under this discretization, a sequence of images $\{\mathbf{u}^n\}$, where $\mathbf{u}^n := \mathbf{u}(t_n, \mathbf{x})$, constitutes a discrete Eulerian description of the field---tracking the evolution of each physical quantity at fixed pixel locations over time. The temporal derivative $\partial_t \mathbf{u}(t_n, \mathbf{x}_{i,j})$ can then be approximated via finite differences:
\begin{equation}
\partial_t \mathbf{u}(t_n, \mathbf{x}_{i,j}) \approx \frac{\mathbf{u}^{n+1}_{i,j} - \mathbf{u}^n_{i,j}}{\Delta t},
\end{equation}
providing a target for learning the spatial operator $\mathcal{D}(\mathbf{u}^n)$ using local convolutional stencils. This setup allows the convolutional neural network to approximate the Eulerian dynamics directly from temporally indexed pixel data.

The standard convolution operation in CNNs is defined in a fixed Euclidean neighborhood on \(\mathcal{G}\) where the relative positions among the kernel elements are fixed and uniform across the Euclidean grid. In this case, the \textit{receptive field} around each pixel $\mathbf{x}_i$---the region of the domain $\Omega$ where a stimulus (i.e., pixel value) can influence the output of the neuron at the pixel---is a fixed, uniformly‐sized window of width $w$. Assuming the square-shaped convolution kernel with the same width and height as in the usual implementation of CNNs, the receptive field $\tilde{\mathcal{R}}_i$ of the node $\mathbf{x}_i$ is defined as a set:
\begin{equation}
  \tilde{\mathcal{R}}_i \;=\; \bigl\{\,\mathbf{x}_j \, \left\vert \, \lVert \mathbf{x}_i - \mathbf{x}_j\rVert_\infty \leq \tfrac{w}{2}\right.\bigr\},
\end{equation}
where $\| \cdot \|_\infty$ is the $l_\infty$ norm. 

This fixed receptive field is where the differential operator $\mathcal{D}$ is modeled. In other words, a CNN built upon the standard convolution operation would attempt to predict $\mathcal{D}(\mathbf{u})$ at $\mathbf{x}_i$ as a combination of function values $\mathbf{u}_j$ sampled from the fixed neighbors $\mathbf{x}_j \in \tilde{\mathcal{R}}_i$.

When the function $u$ is smooth and near-linear, this fixed-neighborhood discretization may yield a faithful approximation of the operator $\mathcal{D}$. However, for $u$ that exhibits strong nonlinearity, sharp gradients, bifurcations, or discontinuities, such a fixed-neighborhood discretization may render approximation errors and numerical instabilities, as already well known in the existing body of numerical simulation and modeling literature. Instead, we would like to adaptively relocate kernel elements (cells) depending on the local function shape of $u$ so that the approximation of $\mathcal{D}$ is more accurate for strong nonlinear cases.

To do so, we introduce \textit{deformable convolutions} that can adaptively relocate cells depending on the function shape of $\mathbf{u}$. The key idea here is to adjust the cells $\mathbf{x}_j$ in $\tilde{\mathcal{R}}_i$ in a Lagrangian manner through shifting their locations by $\Delta \mathbf{x}_j$, based on the complexity of the function shape $\mathbf{u}$ near $\mathbf{x}_i$. The offsets $\Delta x_j$ for each sampling point $j$ relative to the central pixel $i$ are learned by a separate convolutional layer operating on the input $\mathbf{u}(t, x_k)$ for $k\in\mathcal{N}_i$ (the neighborhood of $i$):

 \begin{equation}
    \Delta \mathbf{x}_j
    = h_{ij}\bigl(\textbf{u}(t, \textbf{x}_{k\in\mathcal{N}_i})\bigr)
 \end{equation}

where $h_{ij}$ represents the learned transformation that generates the specific offset for the $j$-th point in the kernel from the input features around $\mathbf{x_i}$. These offsets allow the receptive field to dynamically deform cells and focus on regions most relevant for the current prediction. This adaptive relocation of cells, based on the local function shape of $u$, aims to improve the approximation of $\mathcal{D}$, particularly in scenarios with strong strong nonlinear cases.

\subsubsection{Data Availability}

All datasets used in this study, including those for Burgers' equation and the Navier–Stokes equation, are publicly available via Zenodo at \url{https://zenodo.org/records/13909869}.

Each dataset originates from direct numerical simulations (DNS) that compute the high-strain, nonlinear dynamics governed by the respective physical relations. The DNS data are defined on a fixed Eulerian grid \( (H \times W) \), providing the ground truth for our comparative analyses. The specific mathematical formulations and simulation setups for the flow computations are detailed in Supplementary~\ref{app:appendix_burgers}, \ref{app:appendix_ns}, and \ref{app:appendix_em}.

\subsubsection{Training and Evaluation}

Training procedure, including optimizer configuration, learning rates, batch sizes, and loss functions follows~\citep{nguyen2024parcv2}. We evaluate field-level accuracy using root mean square error (RMSE). For energetic materials, we additionally compute Intersection over Union (IoU), Dice scores, and IoU-weighted RMSE for hotspot geometry (T $>$ 875 K). Detailed metric definitions are provided in Supplementary~\ref{app:hotspot_metrics}.

\subsubsection{Code Availability}

All code supporting this work is openly available on GitHub at \url{https://github.com/baeklab/PARCtorch/tree/main?tab=readme-ov-file}.

\bibliographystyle{unsrt} 
\bibliography{ref} 

\begin{thebibliography}{10}

\bibitem{DBLP:journals/corr/abs-2001-08361}
Jared Kaplan, Sam McCandlish, Tom Henighan, Tom~B. Brown, Benjamin Chess, Rewon Child, Scott Gray, Alec Radford, Jeffrey Wu, and Dario Amodei.
\newblock Scaling laws for neural language models.
\newblock {\em CoRR}, abs/2001.08361, 2020.

\bibitem{DBLP:journals/corr/abs-1905-11946}
Mingxing Tan and Quoc~V. Le.
\newblock Efficientnet: Rethinking model scaling for convolutional neural networks.
\newblock {\em CoRR}, abs/1905.11946, 2019.

\bibitem{Sevilla_2022}
Jaime Sevilla, Lennart Heim, Anson Ho, Tamay Besiroglu, Marius Hobbhahn, and Pablo Villalobos.
\newblock Compute trends across three eras of machine learning.
\newblock In {\em 2022 International Joint Conference on Neural Networks (IJCNN)}, page 1–8. IEEE, July 2022.

\bibitem{sanchez2020learning}
Alvaro Sanchez-Gonzalez, Jonathan Godwin, Tobias Pfaff, Rex Ying, Jure Leskovec, and Peter Battaglia.
\newblock Learning to simulate complex physics with graph networks.
\newblock In {\em International conference on machine learning}, pages 8459--8468. PMLR, 2020.

\bibitem{laney1998computational}
Culbert~B Laney.
\newblock {\em Computational gasdynamics}.
\newblock Cambridge university press, 1998.

\bibitem{SHU1988439}
Chi-Wang Shu and Stanley Osher.
\newblock Efficient implementation of essentially non-oscillatory shock-capturing schemes.
\newblock {\em Journal of Computational Physics}, 77(2):439--471, 1988.

\bibitem{johnson2009numerical}
Claes Johnson.
\newblock {\em Numerical solution of partial differential equations by the finite element method}.
\newblock Courier Corporation, 2009.

\bibitem{osti_4769185}
Francis~H Harlow.
\newblock The particle-in-cell method for numerical solution of problems in fluid dynamics.
\newblock Technical report, Los Alamos National Laboratory (LANL), Los Alamos, NM (United States), 03 1962.

\bibitem{brackbill1988flip}
Jeremiah~U Brackbill, Douglas~B Kothe, and Hans~M Ruppel.
\newblock Flip: a low-dissipation, particle-in-cell method for fluid flow.
\newblock {\em Computer Physics Communications}, 48(1):25--38, 1988.

\bibitem{sulsky1994particle}
Deborah Sulsky, Zhen Chen, and Howard~L Schreyer.
\newblock A particle method for history-dependent materials.
\newblock {\em Computer methods in applied mechanics and engineering}, 118(1-2):179--196, 1994.

\bibitem{bardenhagen2004generalized}
Scott~G Bardenhagen, Edward~M Kober, et~al.
\newblock The generalized interpolation material point method.
\newblock {\em Computer Modeling in Engineering and Sciences}, 5(6):477--496, 2004.

\bibitem{Eiris2023}
A.~Eirís, L.~Ramírez, I.~Couceiro, and et~al.
\newblock Mls-sph-ale: A review of meshless-fv methods and a unifying formulation for particle discretizations.
\newblock {\em Archives of Computational Methods in Engineering}, 30:4959--4981, 2023.

\bibitem{dai2017deformable}
Jifeng Dai, Haozhi Qi, Yuwen Xiong, Yi~Li, Guodong Zhang, Han Hu, and Yichen Wei.
\newblock Deformable convolutional networks.
\newblock In {\em Proceedings of the IEEE international conference on computer vision}, pages 764--773, 2017.

\bibitem{nguyen2024parcv2}
Phong~CH Nguyen, Xinlun Cheng, Shahab Arfaza, Pradeep Seshadri, Yen~T Nguyen, Munho Kim, Sanghun Choi, HS~Udaykumar, and Stephen Baek.
\newblock Parcv2: Physics-aware recurrent convolutional neural networks for spatiotemporal dynamics modeling.
\newblock {\em arXiv preprint arXiv:2402.12503}, 2024.

\bibitem{cheng2024physics}
Xinlun Cheng, Phong~CH Nguyen, Pradeep~K Seshadri, Mayank Verma, Zo{\"e}~J Gray, Jack~T Beerman, HS~Udaykumar, and Stephen~S Baek.
\newblock Physics-aware recurrent convolutional neural networks for modeling multiphase compressible flows.
\newblock {\em International Journal of Multiphase Flow}, page 104877, 2024.

\bibitem{nguyen2022multi}
Yen Nguyen, Pradeep Seshadri, Oishik Sen, D~Barrett Hardin, Christopher~D Molek, and HS~Udaykumar.
\newblock Multi-scale modeling of shock initiation of a pressed energetic material i: The effect of void shapes on energy localization.
\newblock {\em Journal of Applied Physics}, 131(5), 2022.

\bibitem{welle2014microstructural}
EJ~Welle, CD~Molek, RR~Wixom, and P~Samuels.
\newblock Microstructural effects on the ignition behavior of hmx.
\newblock In {\em Journal of Physics: Conference Series}, volume 500, page 052049. IOP Publishing, 2014.

\bibitem{song2021improved}
Jae-Uk Song and Hyun-Gyu Kim.
\newblock An improved material point method using moving least square shape functions.
\newblock {\em Computational Particle Mechanics}, 8(4):751--766, 2021.

\bibitem{toro2013riemann}
Eleuterio~F Toro.
\newblock {\em Riemann solvers and numerical methods for fluid dynamics: a practical introduction}.
\newblock Springer Science \& Business Media, 2013.

\bibitem{luo2016understanding}
Wenjie Luo, Yujia Li, Raquel Urtasun, and Richard Zemel.
\newblock Understanding the effective receptive field in deep convolutional neural networks.
\newblock {\em Advances in neural information processing systems}, 29, 2016.

\bibitem{chen1995universal}
Tianping Chen and Hong Chen.
\newblock Universal approximation to nonlinear operators by neural networks with arbitrary activation functions and its application to dynamical systems.
\newblock {\em IEEE transactions on neural networks}, 6(4):911--917, 1995.

\bibitem{10.1063/1.4938581}
Nirmal~K. Rai and H.~S. Udaykumar.
\newblock Mesoscale simulation of reactive pressed energetic materials under shock loading.
\newblock {\em Journal of Applied Physics}, 118(24):245905, 12 2015.

\bibitem{PhysRevFluids.2.043202}
Nirmal~Kumar Rai, Martin~J. Schmidt, and H.~S. Udaykumar.
\newblock High-resolution simulations of cylindrical void collapse in energetic materials: Effect of primary and secondary collapse on initiation thresholds.
\newblock {\em Phys. Rev. Fluids}, 2:043202, Apr 2017.

\end{thebibliography}

\appendix 
\section{Supplementary}

\begin{figure}[ht!]
    \centering
    \includegraphics[width=\linewidth]{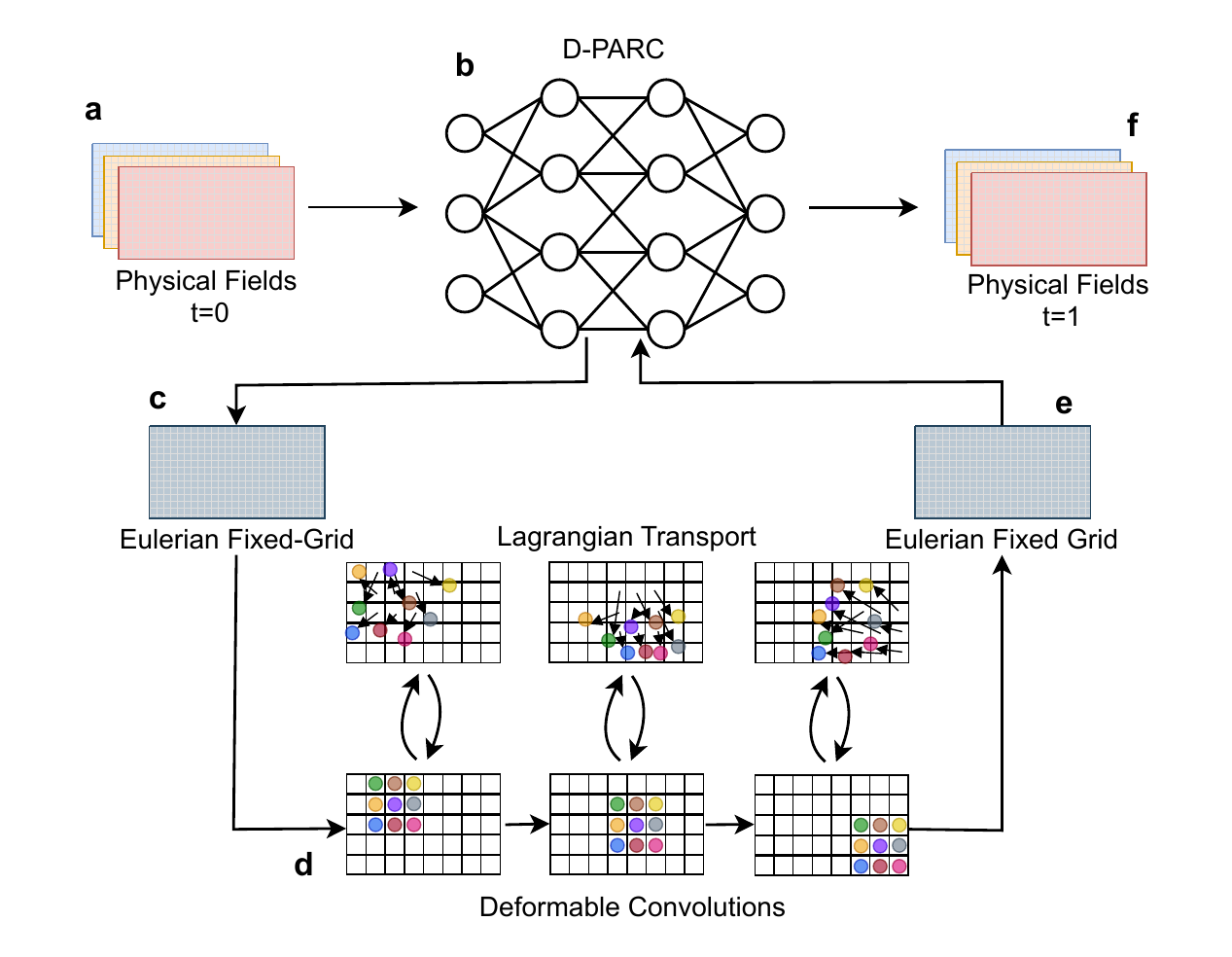}
    \caption{\textbf{D-PARC architecture integrates deformable convolutions with physics-aware recurrent learning.} Diagram of D-PARC's hybrid Lagrangian-Eulerian approach for temporal field prediction. \textbf{a}, Input physical fields at $t=0$ on Eulerian grid (multiple state variables stacked as channels). \textbf{b}, Recurrent neural network processes fields through learned physics-aware operators. \textbf{c}, Feature maps determine adaptive sampling locations based on local flow dynamics, decoupled from main computation path. \textbf{d}, Deformable convolution kernels sample at learned offset positions across the domain. Colored circles represent individual kernel elements displaced from fixed grid positions (top row) to adaptive locations (bottom row); arrows indicate offset vectors. Each spatial location has a unique kernel configuration. \textbf{e}, Bilinear interpolation at offset locations followed by learned weights produces predicted fields projected back onto Eulerian grid. \textbf{f}, Output fields at $t=1$ maintain structured grid representation for next timestep. The architecture enables Lagrangian-type kernel transport (panel d) while maintaining computational efficiency of Eulerian representations (panels a, e, f), combining adaptive sampling with fixed-grid structure.}
\label{fig:DPARC_arch}
\end{figure}

\subsection{Strain-rate}
\label{app:strain_rate}

Specifically, the strain-rate tensor $\mathbf{D}$ used here is built upon the concept of infinitesimal strain-rate in Eulerian continuum mechanics, and is defined as the symmetric part of the velocity gradient tensor. For a two-dimensional velocity field $\mathbf{u} = (u, v)^\top$, $\mathbf{D}$ is formulated as:

\begin{equation*}
    \mathbf{D} = \frac{1}{2}\left[\nabla \mathbf{u} + (\nabla \mathbf{u})^\top\right] =
    \begin{bmatrix}
        \dfrac{\partial u}{\partial x} & \dfrac{1}{2}\left(\dfrac{\partial u}{\partial y} + \dfrac{\partial v}{\partial x}\right) \\[6pt]
        \dfrac{1}{2}\left(\dfrac{\partial u}{\partial y} + \dfrac{\partial v}{\partial x}\right) & \dfrac{\partial v}{\partial y}
    \end{bmatrix}.
\end{equation*}

The diagonal terms, $D_{xx}$ and $D_{yy}$, correspond to normal (extensional or compressional) strain rates along the coordinate directions, whereas the off-diagonal terms represent shear deformation. In the D-PARC analysis, these quantities are evaluated at each pixel from the predicted velocity fields, allowing comparison between local strain regimes and the corresponding kernel responses. Notably, as shown in the main results, kernel behavior differs markedly between high- and low-strain regions.

\subsection{Data Description}
\subsubsection{Burgers' equation}
\label{app:appendix_burgers}

The Burgers' equation is a canonical non-linear partial differential equation that models the interplay between advection and diffusion. Its fundamental balance between the convective term, which leads to wave steepening and shock formation, and the viscous term, which diffuses sharp gradients, provides a non-trivial yet well-understood test case. For the D-PARC model, it serves as a crucial benchmark to evaluate the model's ability to capture and adapt to spatially varying strain-rate fields characteristic of nonlinear fluid flows.

The two-dimensional viscous Burgers’ equation is given in vector form by:
\begin{equation*}
    \frac{\partial \mathbf{u}}{\partial t} + (\mathbf{u} \cdot \nabla) \mathbf{u} = \nu \nabla^2 \mathbf{u}
\end{equation*}

where $\mathbf{u}(\mathbf{r}, t) = [u(\mathbf{r}, t), v(\mathbf{r}, t)]^\top$ is the velocity field at position $\mathbf{r}$ and time $t$, and $\nu$ is the kinematic viscosity. $(\mathbf{u} \cdot \nabla) \mathbf{u}$ is the nonlinear convective (advection) term and $\Delta \mathbf{u}$ is the Laplacian of the velocity field representing viscous diffusion.

The initial velocity field was prescribed using:
\begin{equation*}
    u(\mathbf{r}, t)\big|_{t=0} = v(\mathbf{r}, t)\big|_{t=0} = a \exp\left(-\frac{\|\mathbf{r}\|_2^2}{w}\right)
\end{equation*}

Across the different cases, the parameters used to control the flow field were the initial velocity distribution ($a$,$w$), and the quantity $1/\nu$ which controls the ratio of effects of the inertial and viscous forces.

\begin{table}[!ht]
\centering
\renewcommand{\arraystretch}{1.2}
\setlength{\tabcolsep}{10pt}
\newcolumntype{Y}{>{\centering\arraybackslash}X} 

\begin{tabularx}{0.85\textwidth}{lYY}
\hline
\textbf{Parameter} & \textbf{Training} & \textbf{Testing} \\
\hline
$1/\nu$ (s/cm$^2$) & 
1000, 2500, 5000, 7500, 10000 & 
100, 500, 3000, 6500, 12500, 15000 \\
$a$ (cm/s) & 
0.5, 0.6, 0.7, 0.8, 0.9 & 
0.35, 0.40, 0.45, 0.55, 0.65, 0.75, 0.85, 0.95, 1.00 \\
$w$ (cm) & 
0.7, 0.8, 0.9, 1.0 & 
0.55, 0.6, 0.65, 0.75, 0.85, 0.95, 1.05 \\
\hline
\end{tabularx}
\caption{Training and testing parameters for Burgers used in this study (adopted from~\cite{nguyen2024parcv2}).}
\label{tab:train_test_Raw}
\end{table}

\subsubsection{Navier-Stokes}
\label{app:appendix_ns}

To assess the D-PARC framework on a more physically comprehensive system, we use the incompressible Navier-Stokes equations to model viscous flow past a circular cylinder. This problem is a canonical benchmark in computational fluid dynamics because it encapsulates several critical phenomena, including boundary layer formation, flow separation, and periodic vortex shedding. It provides a rigorous test for a model's ability to capture complex, unsteady flow patterns and fluid-boundary interactions.
The governing equation set is:
\begin{align*}
\frac{\partial \mathbf{u}}{\partial t} + (\mathbf{u} \cdot \nabla)\mathbf{u} &= -\frac{1}{\rho} \nabla p + \nu \nabla^2 \mathbf{u} \\
\nabla \cdot \mathbf{u} &= 0
\end{align*}
Here, in addition to $\mathbf{u}$, $\mathbf{r}$, $t$, and $\nu$ (quantities present in Burgers' equation), $\rho$ is the density and $p$ is the pressure. 

The setup consists of a circle of 0.25m diameter embedded in a 2 m $\times$ 1 m rectangular domain \cite{nguyen2024parcv2}. The circle representing the cylindrical bluff body was centered vertically in the domain and 0.5 m from the left (inlet) edge. The velocity at inlet was uniformly distributed and maintained as ${u_0}$ = 1 m/s. Further, the dynamic viscosity was fixed at 1 kg/ms, and $\rho$ was varied to effectively vary the quantity $1/\nu$. At high values of $1/\nu$, the flow is steady and laminar, while at lower values the flow becomes unsteady and exhibits the characteristic vortex shedding.

\newcolumntype{Y}{>{\centering\arraybackslash}X}

\begin{table}[ht!]
\centering
\renewcommand{\arraystretch}{1.2}
\setlength{\tabcolsep}{10pt}

\begin{tabularx}{0.85\textwidth}{lYY}
\hline
\textbf{Parameter} & \textbf{Training} & \textbf{Testing} \\
\hline
$1/\nu$ (s/m$^2$) & 
30, 40, 80, 100, 150, 200, 250, 300, 400, & 
20, 60, 140, 350, \\
 & 
450, 500, 600, 650, 700, 800, 850, 900, 950 & 
550, 750, 1000 \\
\hline
\end{tabularx}
\caption{Training and testing conditions for Navier-Stokes used in this study (adopted from~\cite{nguyen2024parcv2}).}
\label{tab:train_test_params}
\end{table}

\subsubsection{Energetic Materials}
\label{app:appendix_em}

The energetic materials (EMs) dataset used in this study was derived from high-fidelity direct numerical simulations (DNS) of heterogeneous microstructural samples of EM subjected to shock loading, as originally presented in \cite{nguyen2024parcv2}. These simulations capture the transient thermomechanical response in the shocked microstructures, particularly the localization of the mechanical energy at the heterogeneity (pore in the EM sample) leading to the formation and evolution of localized high-temperature regions called ``hotspots.'' Accurate prediction of hotspot initiation and subsequent spread of reaction is critical to understanding sensitivity and safety in EM systems.

Each simulation instance provides spatiotemporal fields of the state variables $\mathbf{x} := [T, p, \mu]$, where $T$ denotes temperature, $p$ is pressure, and $\mu$ represents microstructure density, along with the velocity components $\mathbf{u} := [u, v]$. The simulations were performed using the in-house code \textsc{SCMITAR3D} \cite{10.1063/1.4938581,PhysRevFluids.2.043202}, under a fixed incident shock pressure of $p_s = 9.5~\mathrm{GPa}$. Data from 100 simulation runs was used for training the models. Each simulation data consisted of 15 temporal snapshots sampled at uniform intervals of $\delta t = 0.17~\mathrm{ns}$. The simulation set was split as 66 cases for training, 32 for testing, and 34 for validation. The dataset captures a wide range of microstructural configurations, as the shockwave leads to the deformation and collapse of the pore accompanied by rich shock interactions and evolution of reaction, making it a representative benchmark for modeling extreme mechanochemical phenomena in EMs. For further details on the simulation setup and parameters, readers are referred to  \cite{nguyen2024parcv2}.

\begin{figure}
    \centering
    \includegraphics[width=\linewidth]{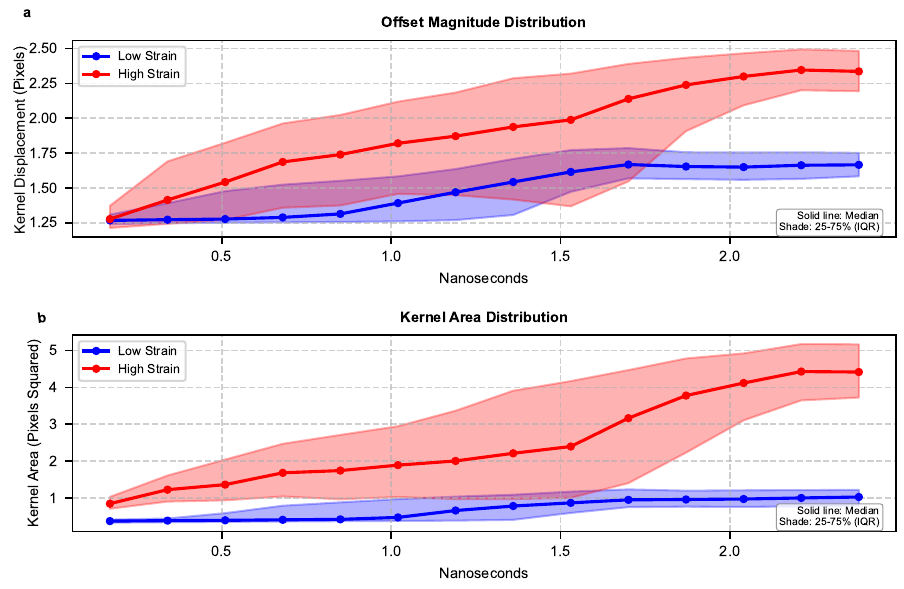}
    \caption{\textbf{Deformable kernels expand receptive fields in high-strain regions of energetic material flow.} Temporal evolution of kernel metrics for EM simulation, stratified by local strain rate regime. Solid lines show median values; shaded regions show interquartile range (25th--75th percentile, IQR). \textbf{a}, Offset magnitude: distance kernel elements displace from fixed grid positions (pixels). High-strain regions (red) show progressively larger offsets as shock-pore interactions intensify; low-strain regions (blue) maintain compact, stable offsets. \textbf{b}, Kernel area: spatial extent measured by convex hull of nine deformed kernel elements (pixels$^2$). High-strain kernels expand substantially (1 to 4.5 pixels$^2$), widening context windows to capture steep gradients. Low-strain kernels remain compact (0.5 to 1.0 pixels$^2$), conserving computational resources where dynamics are smooth. Metrics computed across all spatial locations and timesteps. This quantifies the adaptive behavior visualized in Figure~\ref{fig:spatial_deformable_kernels}, demonstrating individual kernels widen in high-gradient regions while collective scaffolding concentrates multiple kernels at these same locations.}
\label{fig:strain_statistics}
\end{figure}

\subsection{Model Predictions}
\begin{figure}
    \centering
    \includegraphics[width=\linewidth]{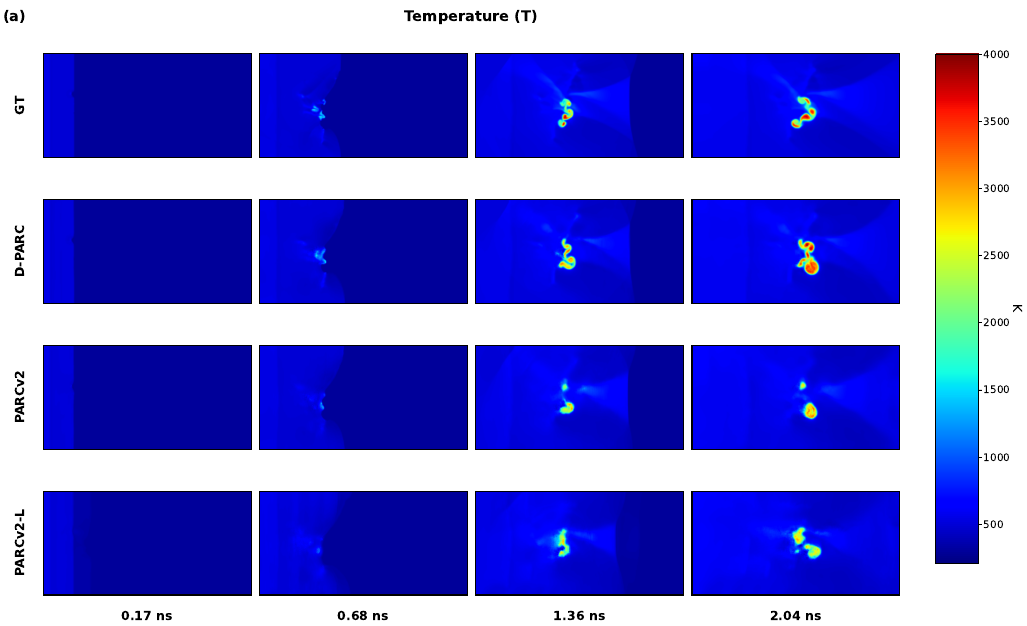}
    \caption{\textbf{D-PARC accurately captures hotspot formation and evolution in energetic material simulations.} Temporal evolution of temperature field (K) during shock-induced pore collapse at four timesteps (0.17, 0.68, 1.36, 2.04 ns). Rows compare ground truth (GT), D-PARC, PARCv2, and PARCv2-L predictions. Hotspots (T $>$ 875 K, yellow-red regions) emerge from shock-pore interactions and grow over time. D-PARC closely matches GT in hotspot location, morphology, and intensity. PARCv2 captures general structure but underresolves peak temperatures. PARCv2-L shows greater deviation in hotspot shape and spatial distribution despite 12.6$\times$ more parameters than D-PARC. Accurate hotspot prediction is critical for shock-to-detonation transition modeling (SDT).}
\label{fig:em_temperature_comparison}
\end{figure}

\begin{figure}
\centering
\includegraphics[width=\linewidth]{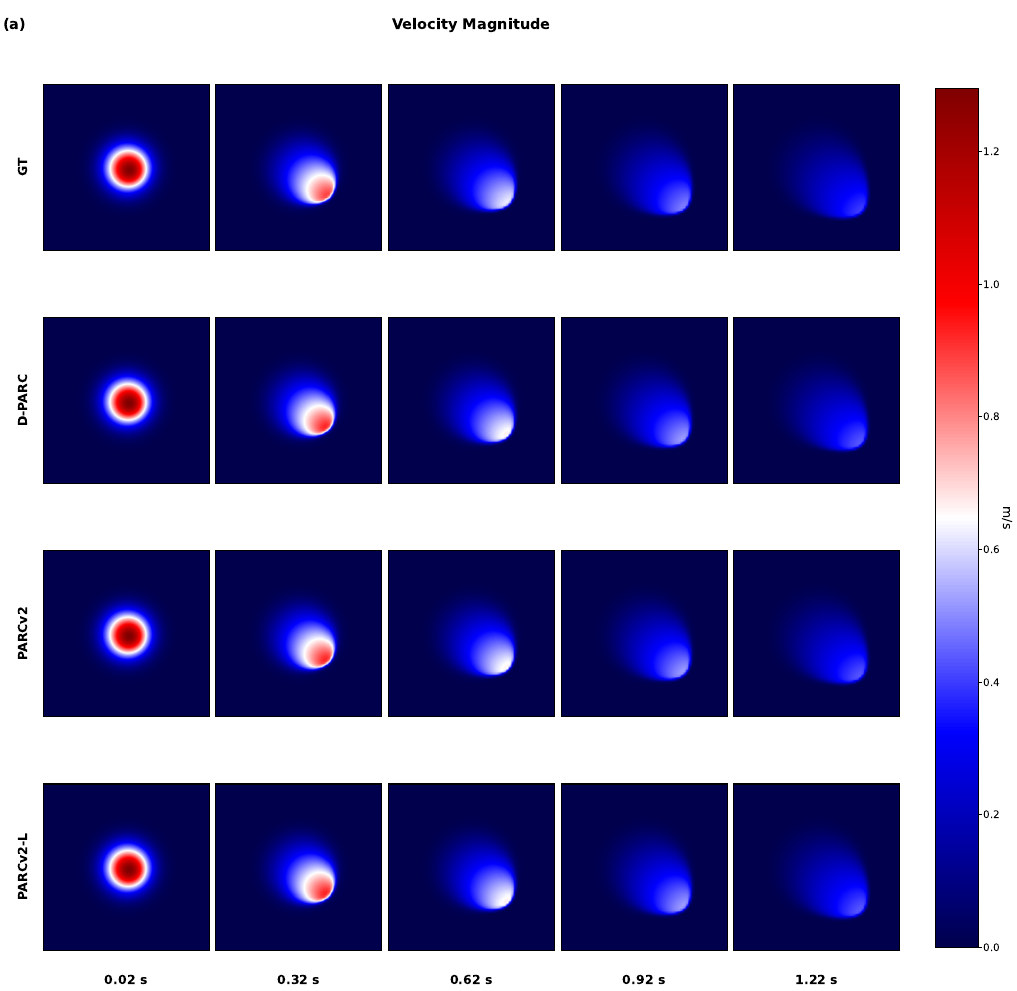}
\caption{\textbf{All models capture shock propagation in Burgers' equation with comparable accuracy.} Temporal evolution of velocity magnitude (m/s) for 2D inviscid compressible flow at five timesteps (0.02, 0.37, 0.62, 0.92, 1.22 s). Rows compare ground truth (GT), D-PARC, PARCv2, and PARCv2-L predictions. All models accurately capture shock front position, circular wave propagation, and velocity decay. Visual agreement reflects similar quantitative performance in this relatively simple flow regime (Table~\ref{tab:compact_dataset_metrics}): D-PARC shows marginal 0.36\% RMSE improvement over PARCv2, while PARCv2-L achieves 1.8\% improvement but at 10$\times$ parameter cost. Comparable performance across models indicates Burgers' equation lacks sufficient complexity to demonstrate D-PARC's adaptive sampling advantage, which becomes pronounced in higher-complexity regimes (Navier-Stokes, EM).}
\label{fig:burgers_comparison}
\end{figure}

\begin{figure}
\centering
\includegraphics[width=\linewidth]{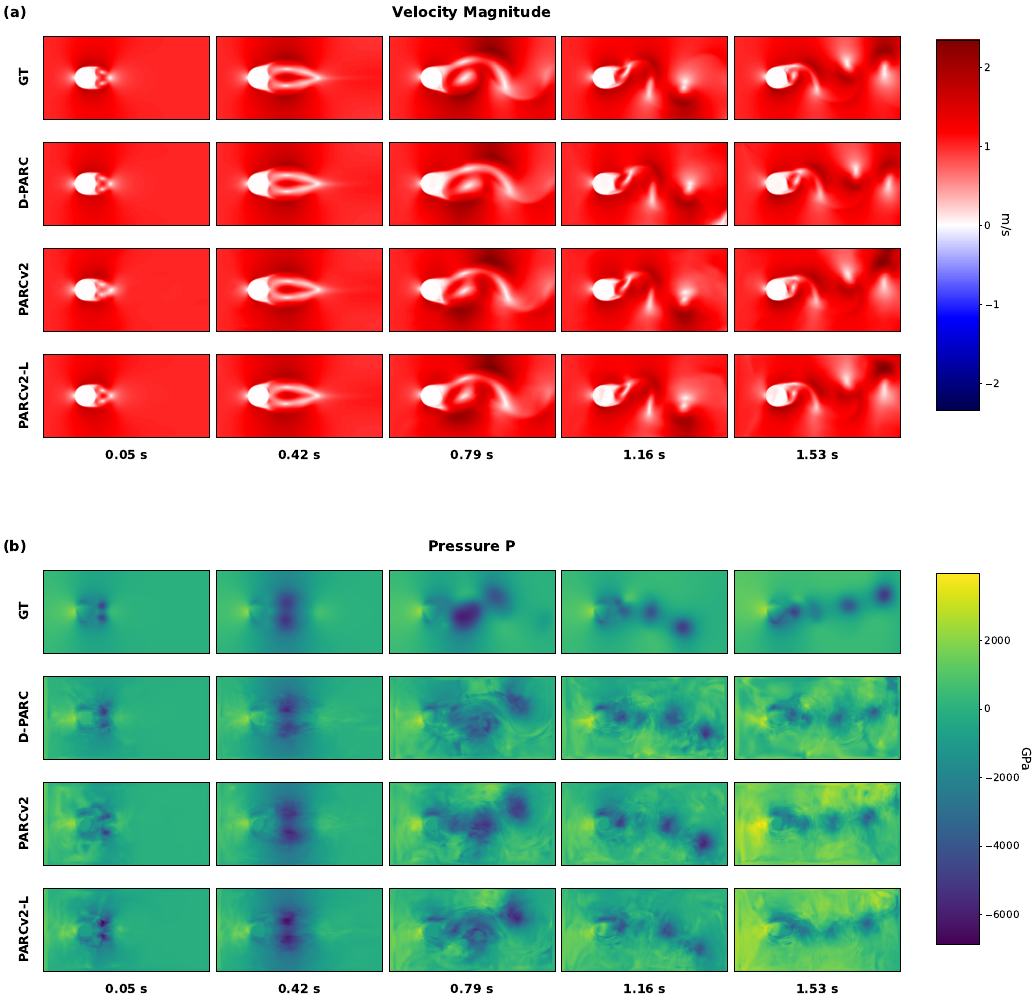}
\caption{\textbf{D-PARC accurately captures vortex dynamics and pressure fields in Navier-Stokes flow.} Temporal evolution of velocity magnitude (top, m/s) and pressure (bottom, Gpa) for incompressible flow over bluff body (Re = 20-1000) at five timesteps (0.05, 0.42, 0.79, 1.16, 1.53 s). Rows compare ground truth (GT), D-PARC, PARCv2, and PARCv2-L predictions. D-PARC accurately resolves vortex shedding patterns, wake structure, and pressure gradients across the cylinder. PARCv2 shows moderate agreement but with smoothed vortical structures. PARCv2-L exhibits increased deviation in wake dynamics and pressure distribution despite 12.6$\times$ more parameters. Quantitative performance (Table~\ref{tab:compact_dataset_metrics}): D-PARC reduces velocity RMSE by 9.61\% and pressure RMSE by 1.56\% versus PARCv2, significantly outperforming PARCv2-L (2.32\% velocity, 0.42\% pressure). Visual differences become pronounced as flow complexity increases relative to Burgers' equation.}
\label{fig:ns_comparison}
\end{figure}

\textbf{Figure~\ref{fig:em_temperature_comparison}} displays the EM predictions of temperature hotspots by D-PARC, PARCv2, and PARCV2-L. \textbf{Figure~\ref{fig:burgers_comparison}} displays the predictions of velocity magnitude for Burgers' equation by D-PARC, PARCv2, and PARCv2-L. \textbf{Figure~\ref{fig:ns_comparison}} displays the predictions of velocity magnitude and pressure evolution for Navier Stokes by PARCv2, the deeper PARCv2-L, and D-PARC.

\subsection{Effective Receptive Field}
\subsubsection{Burgers' Equation}

The ERF analysis reveals consistent trends to EM and Navier Stokes. However, D-PARC achieved only a \(0.36\%\) RMSE reduction over the baseline for Burgers' equation which was the smallest improvement across all three datasets. The pattern suggests that for the relatively simpler shock dynamics in Burgers' equations, the fixed-kernel architecture of PARCv2 is already well suited to capture the dominant features. The problem exhibits predictable shock propagation without material deformation or multiphysics coupling. PARCv2's uniform sampling strategy proves sufficiently efficient for this regime, explaining the reason D-PARC yields only marginal improvement. 

\begin{table}[!ht]
\centering
\small
\begin{tabular}{lcccccc}
\toprule
\multirow{2}{*}{\textbf{Thresh.}} & \multicolumn{2}{c}{\textbf{D-PARC}} & \multicolumn{2}{c}{\textbf{PARCv2}} & \multicolumn{2}{c}{\textbf{PARCv2-L}} \\
\cmidrule(lr){2-3} \cmidrule(lr){4-5} \cmidrule(lr){6-7}
\textbf{(\% of max)} & \textbf{Area} & \textbf{Sparsity (\%)} & \textbf{Area} & \textbf{Sparsity (\%)} & \textbf{Area} & \textbf{Sparsity (\%)} \\
\midrule
\multicolumn{7}{l}{\textit{Theoretical RF: PARCv2 = 324 px, D-PARC/PARCV2-L = 8,100 px}} \\
\midrule
\multicolumn{7}{l}{\textbf{High-Gradient (Shock) Regions} (n=3, mean±std)} \\
\midrule
0.5  & \textbf{866±69}   & \textbf{78.85±1.69} & 561±44   & 0.00±0.00   & 3030±380  & 26.03±9.27 \\
1.0  & \textbf{608±75}   & \textbf{85.15±1.83} & 522±41   & 0.00±0.00   & 2495±442  & 39.10±10.80 \\
2.0  & \textbf{379±51}   & \textbf{90.75±1.24} & 468±37   & 0.00±0.00   & 1891±408  & 53.82±9.96 \\
3.0  & \textbf{297±50}   & \textbf{92.75±1.23} & 431±40   & 0.00±0.00   & 1545±385  & 62.28±9.39 \\
5.0  & \textbf{213±49}   & \textbf{94.79±1.20} & 377±38   & 0.00±0.00   & 1109±377  & 72.92±9.21 \\
7.5  & \textbf{149±36}   & \textbf{96.35±0.88} & 326±44   & 5.86±4.24  & 817±360   & 80.06±8.80 \\
10.0 & \textbf{108±28}   & \textbf{97.36±0.67} & 285±51   & 15.43±11.05 & 646±329   & 84.22±8.02 \\
\midrule
\multicolumn{7}{l}{\textbf{Low-Gradient (Smooth) Regions} (n=3, mean±std)} \\
\midrule
0.5  & \textbf{551±29}   & \textbf{86.56±0.70} & 461±22   & 0.00±0.00   & 2689±318  & 34.36±7.76 \\
1.0  & \textbf{499±21}   & \textbf{87.81±0.51} & 422±13   & 0.00±0.00   & 2069±275  & 49.50±6.71 \\
2.0  & \textbf{443±29}   & \textbf{89.18±0.70} & 377±13   & 0.00±0.00   & 1447±224  & 64.66±5.48 \\
3.0  & \textbf{408±29}   & \textbf{90.03±0.71} & 337±20   & 1.34±1.89  & 1085±214  & 73.51±5.22 \\
5.0  & \textbf{348±31}   & \textbf{91.50±0.75} & 275±32   & 15.12±9.96 & 672±118   & 83.59±2.88 \\
7.5  & \textbf{289±31}   & \textbf{92.94±0.75} & 227±44   & 30.04±13.44 & 466±39   & 88.62±0.95 \\
10.0 & \textbf{253±35}   & \textbf{93.82±0.85} & 186±46   & 42.59±14.12 & 347±23   & 91.53±0.55 \\
\bottomrule
\end{tabular}
\caption{\textbf{ERF analysis for Burgers equation shows adaptive selectivity in canonical compressible flow.} Effective receptive field analysis for inviscid compressible flow (Burgers' equation) at $T=47$. Effective Area in pixels (of 8,100 total domain). Theoretical receptive field (TRF): PARCv2 = 324 pixels; D-PARC/PARCv2-L = 8,100 pixels (entire domain). High-gradient regions ($n=3$): shock fronts with steep velocity gradients. Low-gradient regions ($n=3$): smooth rarefaction zones. D-PARC achieves highest sparsity across all thresholds in both regimes. PARCv2 shows zero sparsity (effective area exceeds TRF capacity), indicating it utilizes its entire available context window without selective filtering.}
\label{tab:erf_threshold_sensitivity_burgers}
\end{table}

\begin{figure}
    \centering
    \includegraphics[width=0.9\linewidth]{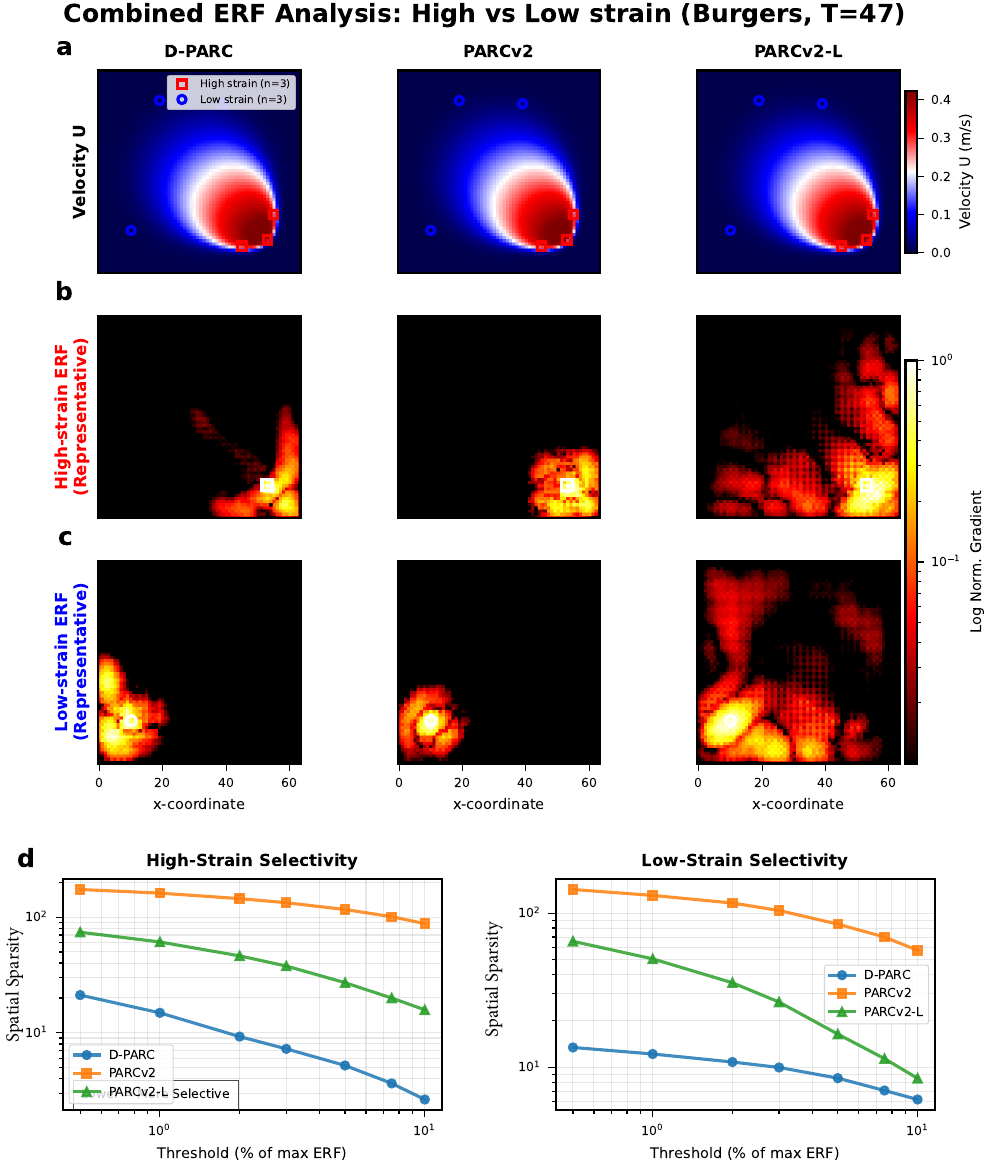}
    \caption{\textbf{ERF visualization for Burgers equation demonstrates selective filtering in 2D compressible flow.} Effective receptive field analysis for Burgers' equation at $T=47$, comparing D-PARC, PARCv2, and PARCv2-L. \textbf{a}, Velocity magnitude field (m/s) with sampled locations: high-gradient regions (red squares, $n=3$) at shock fronts; low-gradient regions (white circles, $n=3$) in smooth zones. \textbf{b}, Representative high-gradient ERF heatmaps at $\tau=1\%$. D-PARC shows compact, focused influence; PARCv2-L displays broader diffusion. \textbf{c}, Representative low-gradient ERF heatmaps at $\tau=1\%$. D-PARC maintains selectivity; PARCv2-L expands unnecessarily. \textbf{d}, Sparsity curves (log-log scale) across thresholds (0.5--10\%). D-PARC achieves lowest utilization in both regimes. See Table~\ref{tab:erf_threshold_sensitivity_burgers} for complete metrics.}
\label{fig:erf_burgers}
\end{figure}

\subsubsection{Navier-Stokes}

\begin{figure}
    \centering
    \includegraphics[width=\linewidth]{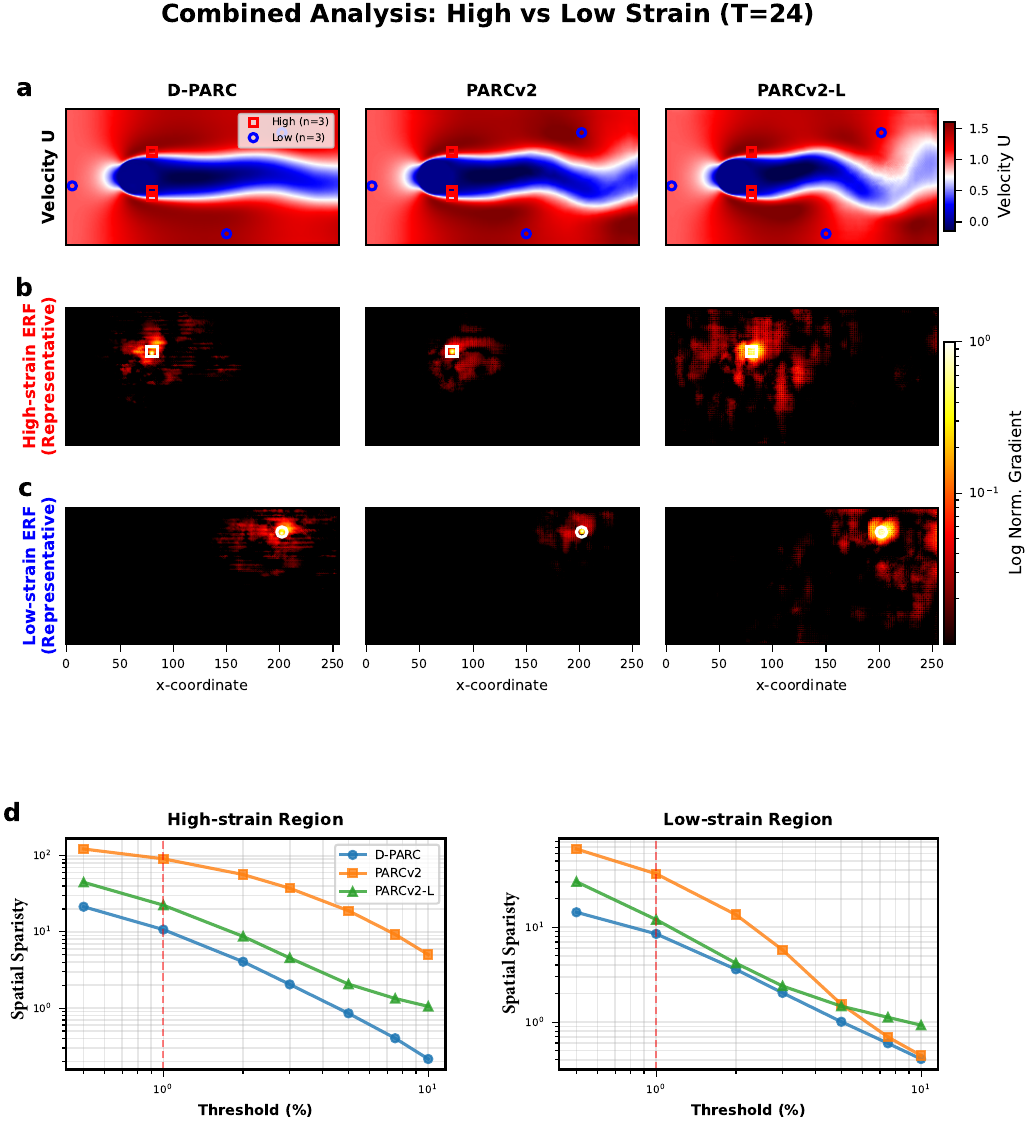}
    \caption{\textbf{ERF visualization for Navier-Stokes flow demonstrates selective filtering in incompressible viscous flow.} Effective receptive field analysis for incompressible flow over bluff body (Re = 20-1000) at $T=24$, comparing D-PARC, PARCv2, and PARCv2-L. \textbf{a}, Velocity magnitude field (m/s) with sampled locations: high-strain regions (red squares, $n=3$) at vortex cores and shear layers; low-strain regions (white circles, $n=3$) in far-field flow. \textbf{b}, Representative high-strain ERF heatmaps at $\tau=1\%$. D-PARC shows compact, focused influence; PARCv2-L displays broader, less selective diffusion. \textbf{c}, Representative low-strain ERF heatmaps at $\tau=1\%$. D-PARC contracts dramatically in smooth regions; PARCv2-L maintains unnecessary broad context. \textbf{d}, Sparsity curves (log-log scale) across thresholds (0.5--10\%). D-PARC achieves lowest utilization (highest selectivity) in both regimes. See Table~\ref{tab:erf_threshold_sensitivity_navierstokes} for complete metrics.}
\label{fig:erf_ns}
\end{figure}

The Navier-Stokes ERF analysis (Table~\ref{tab:erf_threshold_sensitivity_navierstokes} and Figure~\ref{fig:erf_ns}) reveals patterns that depicts D-PARC's substantial performance improvements of \(9.6\%\) and \(1.5\%\) in velocity magnitude and pressure compared to the baseline (Table~\ref{tab:compact_dataset_metrics}). These gains also significantly exceed those of PARCv2-L despite that model having \(14x\) more parameters. 

In high-strain regions, D-PARC achieves substantially higher sparsity across all thresholds reaching \(99.79\%\) at \(\tau=10\%\) compared to PARCv2's \(94.96\%\) and PARCv2-L's \(98.95\%\). This demonstrates D-PARC's ability to efficiently concentrate computational resources on complex flow regimes that dominate the global physics. As such the adaptive kernel allows D-PARC to track advecting vortices. 

In low strain regions away from the wake, D-PARC also exhibits higher sparsity compared to PARCv2 and PARCv2-L, indicating that uniform sampling remains efficient where the flow varies smoothly. This selective advantage demonstrates ``active filtration". PARCv2-L's modest improvements despite 14x more parameters underscore that increasing model capacity cannot replicate the benefits of HLE-behavior in PADL models. 

\begin{table}[!ht]
\centering
\small
\begin{tabular}{lcccccc}
\toprule
\multirow{2}{*}{\textbf{Thresh.}} & \multicolumn{2}{c}{\textbf{D-PARC}} & \multicolumn{2}{c}{\textbf{PARCv2}} & \multicolumn{2}{c}{\textbf{PARCv2-L}} \\
\cmidrule(lr){2-3} \cmidrule(lr){4-5} \cmidrule(lr){6-7}
\textbf{(\%)} & \textbf{Area} & \textbf{Spar. (\%)} & \textbf{Area} & \textbf{Spar. (\%)} & \textbf{Area} & \textbf{Spar. (\%)} \\
\midrule
\multicolumn{7}{l}{\textit{Theoretical RF: PARCv2 = 6,084 px, D-PARC/PARCv2-L = 101,124 px }} \\
\midrule
\multicolumn{7}{l}{\textbf{High-Strain Regions} (n=3, mean±std)} \\
\midrule
0.5  & \textbf{7013±244} & \textbf{78.60±0.74} & 7453±1905 & 6.85±9.69 & 14680±1454 & 55.20±4.44 \\
1.0  & \textbf{3513±268} & \textbf{89.28±0.82} & 5538±2235 & 19.63±27.76 & 7347±1034 & 77.58±3.16 \\
2.0  & \textbf{1331±208} & \textbf{95.94±0.63} & 3446±1972 & 43.36±32.41 & 2859±410 & 91.27±1.25 \\
3.0  & \textbf{671±100}  & \textbf{97.95±0.30} & 2274±1557 & 62.63±25.59 & 1490±179 & 95.45±0.55 \\
5.0  & \textbf{279±44}   & \textbf{99.15±0.13} & 1154±936 & 81.04±15.38 & 676±1 & 97.94±0.00 \\
7.5  & \textbf{132±48}   & \textbf{99.60±0.15} & 564±525 & 90.73±8.63 & 438±32 & 98.66±0.10 \\
10.0 & \textbf{70±35}    & \textbf{99.79±0.11} & 306±286 & 94.96±4.70 & 345±27 & 98.95±0.08 \\
\midrule
\multicolumn{7}{l}{\textbf{Low-Strain Regions} (n=3, mean±std)} \\
\midrule
0.5  & \textbf{4720±1124} & \textbf{85.59±3.43} & 4075±165 & 33.02±2.71 & 9909±2838 & 69.76±8.66 \\
1.0  & \textbf{2793±877}  & \textbf{91.48±2.68} & 2229±438 & 63.36±7.20 & 3949±1984 & 87.95±6.05 \\
2.0  & \textbf{1180±427}  & \textbf{96.40±1.30} & 827±267  & 86.41±4.39 & 1371±747 & 95.82±2.28 \\
3.0  & \textbf{666±245}   & \textbf{97.97±0.75} & 351±179  & 94.23±2.93 & 787±331 & 97.60±1.01 \\
5.0  & \textbf{330±125}   & \textbf{98.99±0.38} & 94±77    & 98.46±1.26 & 482±146 & 98.53±0.44 \\
7.5  & \textbf{196±62}    & \textbf{99.40±0.19} & 42±25    & 99.30±0.40 & 368±98 & 98.88±0.30 \\
10.0 & \textbf{134±33}    & \textbf{99.59±0.10} & 27±4     & 99.56±0.07 & 304±75 & 99.07±0.23 \\
\bottomrule
\end{tabular}
\caption{\textbf{ERF analysis for Navier-Stokes dataset shows similar adaptive selectivity patterns.} Effective receptive field analysis for incompressible flow over bluff body (Re = 20-1000) at $T=24$. Theoretical receptive field (TRF): PARCv2 = 6,084 pixels; D-PARC/PARCv2-L = 101,124 pixels. High-strain regions ($n=3$): vortex cores, shear layers, and separation zones. Low-strain regions ($n=3$): upstream and far-field flow. D-PARC achieves highest sparsity in high-strain regions across all thresholds, demonstrating consistent selective behavior.}
\label{tab:erf_threshold_sensitivity_navierstokes}
\end{table}

\subsection{Individual Kernel Analysis}
\label{app:ind_an}
\subsubsection{Burgers}

\begin{figure}
    \centering
    \includegraphics[width=\linewidth]{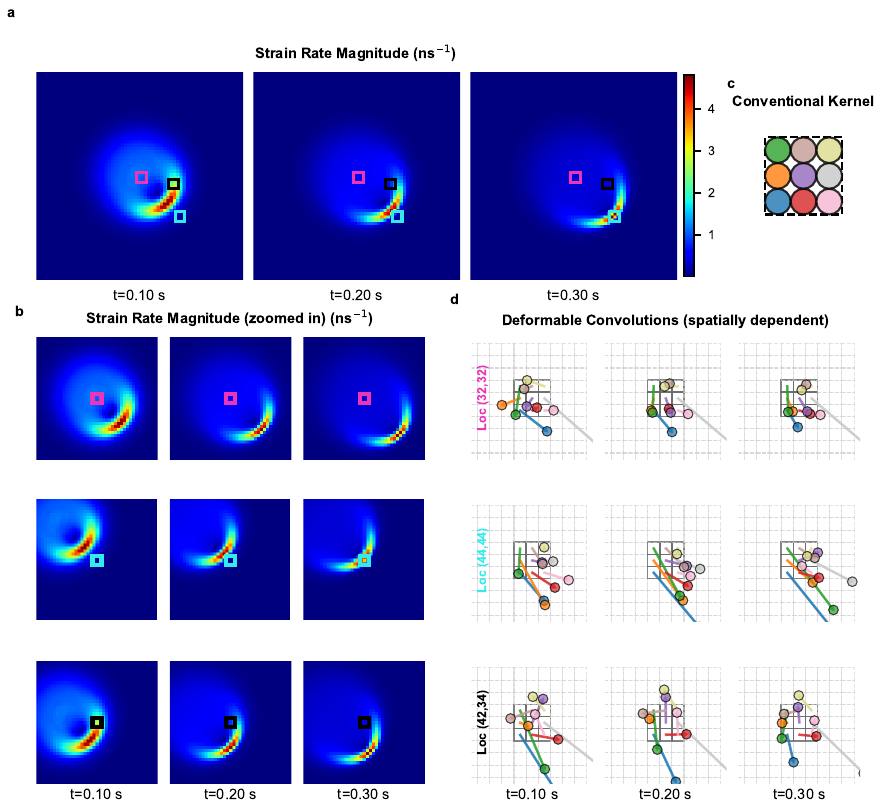}
    \caption{\textbf{Deformable kernels adapt to shock fronts in Burgers' equation.} Visualization of D-PARC kernel behavior during shock formation in 2D inviscid compressible flow. Structure similar to Figure~\ref{fig:spatial_deformable_kernels} but for Burgers dataset. \textbf{a}, Strain rate magnitude field (s$^{-1}$) at three timesteps ($t = 0.1, 0.2, 0.3$ s). Colored squares mark three locations tracking shock development: Loc(pink) behind shock front, Loc(cyan) at shock front, Loc(black) in high-gradient zone. \textbf{b}, Zoomed views (40$\times$40 pixels) showing local strain patterns; white squares indicate 3$\times$3 kernel footprint. \textbf{c}, Conventional kernel with fixed grid positions. \textbf{d}, Deformed kernel configurations across timesteps. Colored circles show element positions after learned offsets; arrows show displacement vectors. Kernels expand substantially at shock fronts where gradients steepen, while remaining compact in smooth regions, demonstrating adaptive context modulation for discontinuity resolution.}
\label{fig:appendix_spatial_deformable_kernels_burgers}
\end{figure}

\begin{figure}
    \centering
    \includegraphics[width=\linewidth]{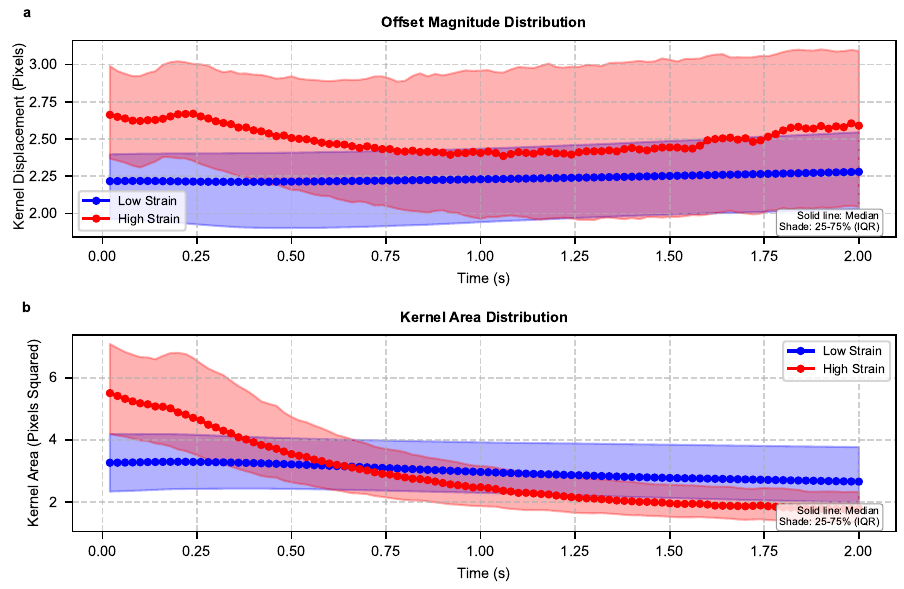}
    \caption{\textbf{Deformable kernels expand receptive fields in high-gradient regions of Burgers' equation.} Temporal evolution of kernel metrics for 2D inviscid compressible flow (Burgers' equation), stratified by local gradient regime. Solid lines show median values; shaded regions show interquartile range (25th--75th percentile, IQR). \textbf{a}, Offset magnitude: distance kernel elements displace from fixed grid positions (pixels). High-gradient regions (red, shock fronts) show larger offsets (~2.5 pixels) than low-gradient regions (blue, smooth zones, ~2.2 pixels). Clear separation persists throughout shock propagation. \textbf{b}, Kernel area: spatial extent measured by convex hull of nine deformed kernel elements (pixels$^2$). High-gradient kernels initially expand substantially (5.5 to 2.0 pixels$^2$) then contract as shock sharpens; low-gradient kernels remain compact and stable (~3.0 pixels$^2$). Decreasing area reflects kernel refinement as discontinuity forms—initially dispersed sampling during shock formation, then convergence as shock becomes sharper. Metrics computed across all spatial locations and timesteps. Consistent with EM and NS datasets (Figures~\ref{fig:strain_statistics},~\ref{fig:ns_strain_statistics}), individual kernels adaptively widen in high-gradient regions, with particularly pronounced temporal evolution at shock discontinuities.}
\label{fig:burgers_strain_statistics}
\end{figure}

Visualizing the individual kernel behavior of D-PARC for Burgers' equation illustrates kernel elements following the dissipation of the shock front. Kernel elements move from their conventional location to the bottom right (\textbf{Figure~\ref{fig:appendix_spatial_deformable_kernels_burgers}}). In areas of high strain, the kernel elements transport larger distance throughout the entire simulation. Similarly, the area of the kernel distribution is larger in high strain areas, until the strain area becomes very small and diffused. Around timestep 30 the kernel area for high strain regions becomes smaller than low strain areas (\textbf{Figure~\ref{fig:burgers_strain_statistics}}).

\subsubsection{Navier-Stokes}

\begin{figure}
    \centering
    \includegraphics[width=\linewidth]{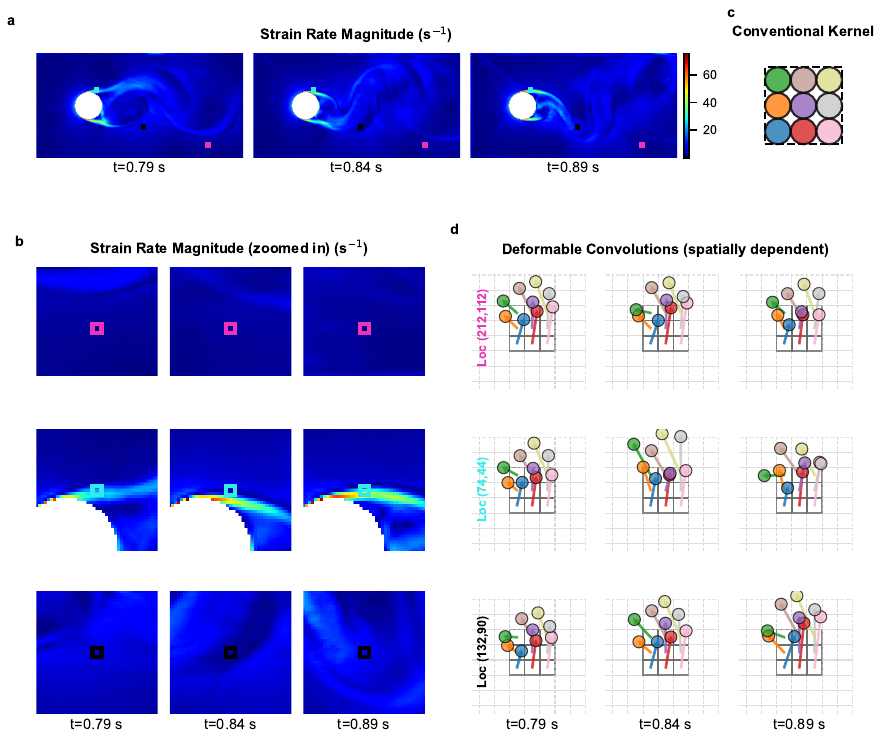}
    \caption{\textbf{Deformable kernels adapt to vortex cores and shear layers in Navier-Stokes flow.} Visualization of D-PARC kernel behavior during vortex shedding in incompressible flow over cylinder (Re = 20-1000). \textbf{a}, Strain rate magnitude field (s$^{-1}$) at three timesteps ($t = 0.79, 0.84, 0.89$ s). Colored squares mark three locations: Loc(pink) away from flow, Loc(cyan) in separating shear layer from cylinder, Loc(black) in far-field flow. \textbf{b}, Zoomed views (40$\times$40 pixels) showing local strain patterns; white squares indicate 3$\times$3 kernel footprint. \textbf{c}, Conventional kernel with fixed grid positions. \textbf{d}, Deformed kernel configurations across timesteps. Colored circles show element positions after learned offsets; arrows show displacement vectors. Kernels expand moderately at shear layers and cylinder boundary while remaining compact in far-field, demonstrating adaptive context modulation for viscous flow features.}
\label{fig:appendix_spatial_deformable_kernels_ns}
\end{figure}

\begin{figure}
    \centering
    \includegraphics[width=\linewidth]{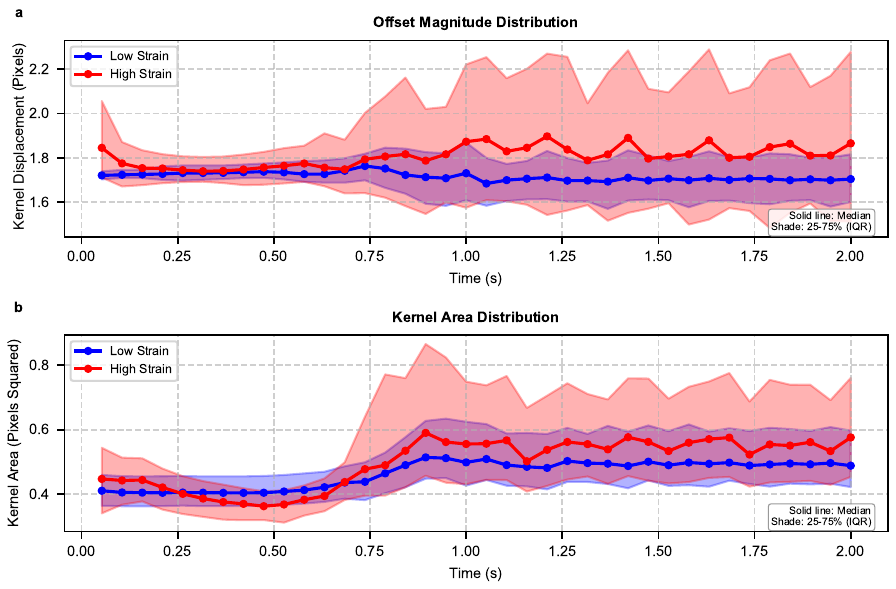}
    \caption{\textbf{Deformable kernels expand receptive fields in high-strain regions of Navier-Stokes flow.} Temporal evolution of kernel metrics for incompressible flow over bluff body (Re = 20-1000), stratified by local strain rate regime. Solid lines show median values; shaded regions show interquartile range (25th--75th percentile, IQR). \textbf{a}, Offset magnitude: distance kernel elements displace from fixed grid positions (pixels). High-strain regions (red, vortex cores and shear layers) show larger offsets (~1.8 pixels) than low-strain regions (blue, far-field flow, ~1.7 pixels). Separation is less pronounced than in EM dataset due to milder gradients in viscous flow. \textbf{b}, Kernel area: spatial extent measured by ellipse fit to nine deformed kernel elements (pixels$^2$). High-strain kernels expand moderately (0.4 to 0.6 pixels$^2$) to capture vortical structures; low-strain kernels remain compact (0.4 to 0.5 pixels$^2$). Metrics computed across all spatial locations and timesteps. Consistent with EM results (Figure~\ref{fig:strain_statistics}), individual kernels adaptively widen in high-gradient regions, though the effect is less dramatic in this lower-Reynolds-number viscous flow regime.}
\label{fig:ns_strain_statistics}
\end{figure}

For the individual kernel behavior of D-PARC on incompressible flow, the kernels exhibit similar behavior to compressible flow. In the figure~\ref{fig:appendix_spatial_deformable_kernels_ns}, the spatial locations of kernel elements examined vary in regions of high and low strain.  Area of high strain experience greater magnitude of transport and large area of the kernel Figure~\ref{fig:ns_strain_statistics}.

\subsection{Multiple Kernel Analysis}
\subsubsection{Burgers' Equation}
\begin{figure}[!ht]
    \centering
    \includegraphics[width=\linewidth]{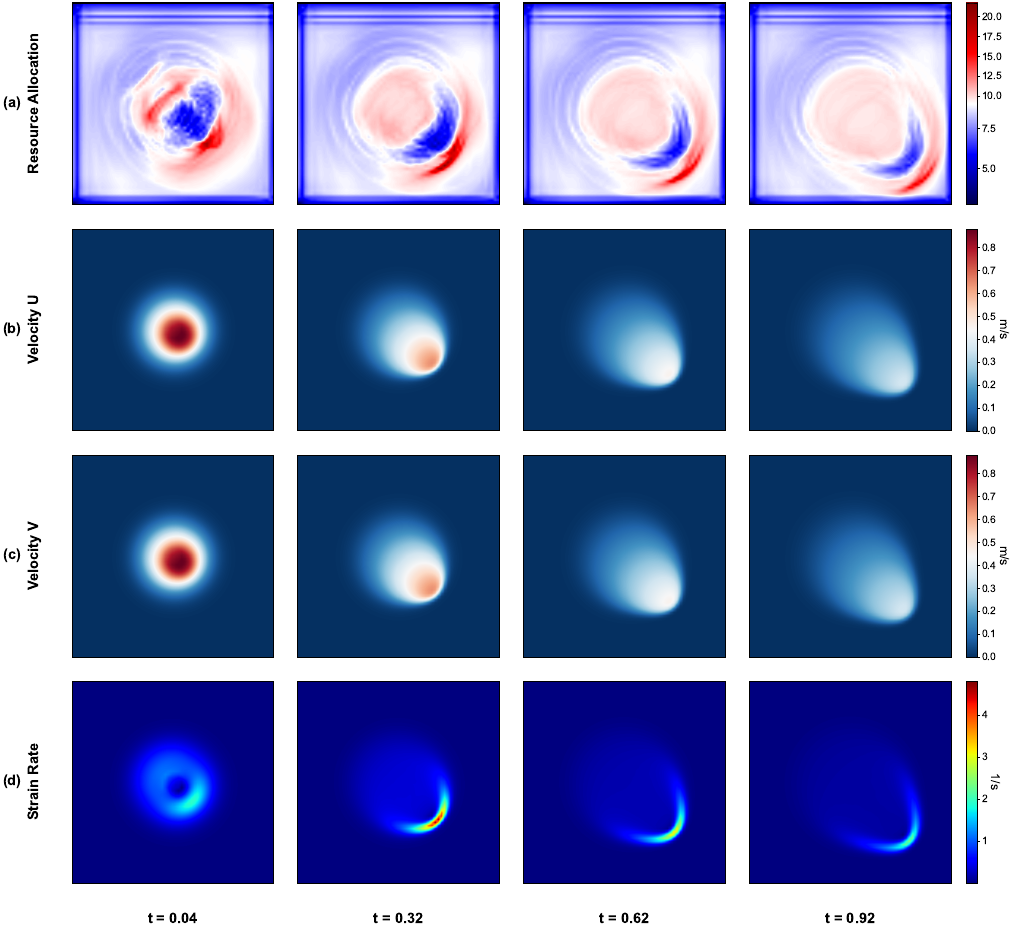}
    \caption{\textbf{D-PARC concentrates resources on shock front formation in Burgers' equation.} Temporal evolution of resource allocation and physical fields for 2D inviscid compressible flow at four timesteps ($t = 0.04, 0.32, 0.62, 0.92$ s). \textbf{a}, Learned resource allocation quantified by cumulative interpolation density. Diverging colormap centered at 9 (baseline sampling); values $>9$ (red/yellow) indicate computational clustering. Resources progressively concentrate on developing shock front as discontinuity forms. \textbf{b}, Horizontal velocity $u$ (m/s) showing initial smooth field steepening into sharp shock. \textbf{c}, Vertical velocity $v$ (m/s) showing complementary shock structure. \textbf{d}, Strain rate magnitude (s$^{-1}$) maximized along shock discontinuity. D-PARC autonomously tracks shock formation, intensifying resource allocation as velocity gradients steepen from smooth initial conditions to fully developed discontinuity.}
\label{fig:appendix_burgers_feature_evolution}
\end{figure}

Figure~\ref{fig:appendix_burgers_feature_evolution} illustrates D-PARC's adaptive resource allocation for Burgers' equation across four timesteps. D-PARC concentrates computational resources at and ahead of the shock front where velocity gradients are steepest. The majority of the domain exhibits significantly reduced sampling compared to the uniform baseline of nine visits per pixel in standard CNNs. 

As the shock propagates across the domain, the resource allocation pattern tracks this motion maintaining concentrated sampling at the moving front. However, the modest performance improvement for Burgers' equation (Table~\ref{tab:compact_dataset_metrics}) suggests that this adaptive strategy provides limited advantage for such simple shock dynamics, consistent with numerical simulation literature that hybrid methods excel primarily for complex flow phenomena. 

\subsubsection{Navier-Stokes}
\begin{figure}[!ht]
    \centering
    \includegraphics[width=\linewidth]{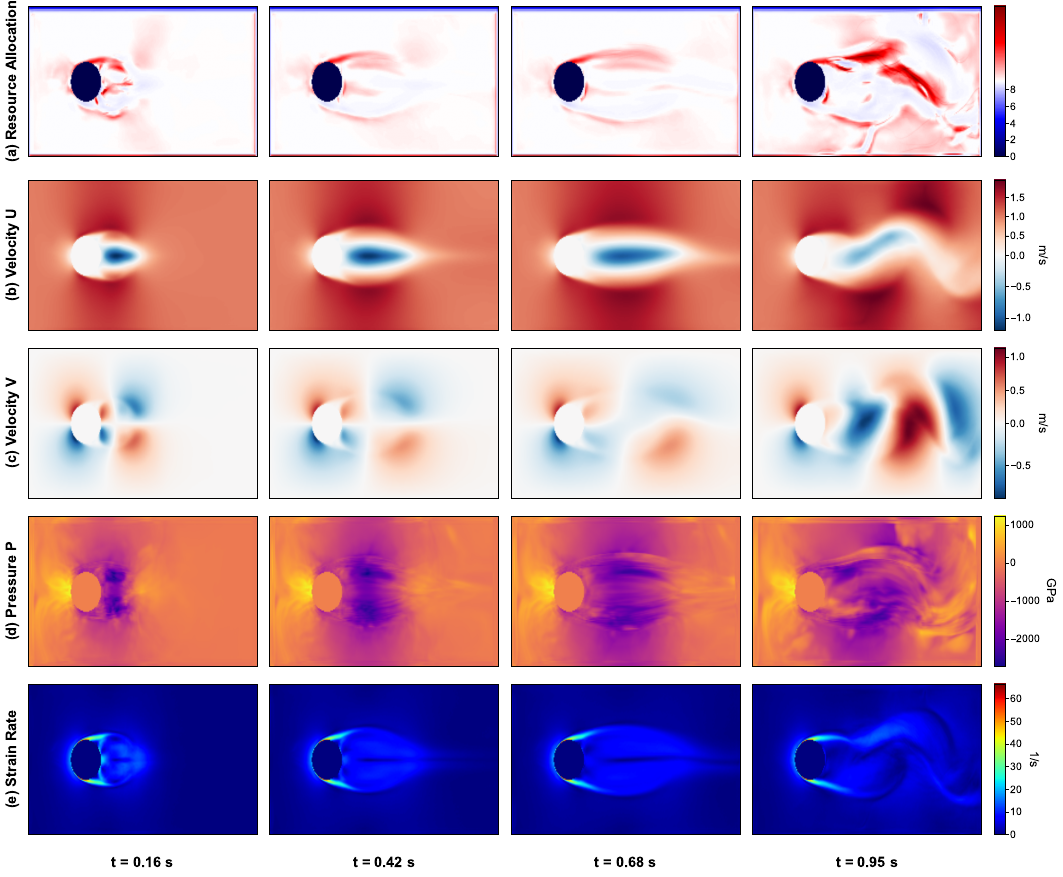}
    \caption{\textbf{D-PARC concentrates resources on vortex shedding and shear layers in Navier-Stokes flow.} Temporal evolution of resource allocation and physical fields for incompressible flow over cylinder (Re = 20-1000) at four timesteps ($t = 0.16, 0.42, 0.68, 0.95$ s). \textbf{a}, Learned resource allocation quantified by cumulative interpolation density. Diverging colormap centered at 9 (baseline sampling); values $>9$ (red) indicate computational clustering. Cylinder interior removed for clarity. \textbf{b}, Horizontal velocity $u$ (m/s). \textbf{c}, Vertical velocity $v$ (m/s). Coolwarm colormaps centered at zero for directional flow. \textbf{d}, Pressure field (GPa) showing high pressure upstream and low pressure in wake. \textbf{e}, Strain rate magnitude (s$^{-1}$) highlighting fluid deformation in shear layers. D-PARC autonomously concentrates resources (panel a) on cylinder boundaries, vortex cores, and separating shear layers—regions with steep velocity gradients and high strain rates. Resource allocation tracks vortex shedding dynamics temporally, intensifying where Kármán vortex street develops.}
\label{fig:appendix_ns_feature_evolution}
\end{figure}

Figure~\ref{fig:appendix_ns_feature_evolution}, illustrates the scaffolding of kernel elements for Navier-Stokes flow across four time steps. The resource allocation map reveals that D-PARC consistently concentrates resources on physically salient features. Conversely, D-PARC significantly reduces sampling sampling in the upstream flow and far regions where velocity and pressure vary smoothly. 

As vortices shed and convect downstream, the resource allocation pattern tracks these structures maintain high sampling density where the flow exhibits complex dynamics. This dynamic reallocation demonstrates Lagrangian-like behavior analogous to PIC methods, where computational resources follow evolving material features. Moreover, the substantial performance improvement for Navier-Stokes (Table~\ref{tab:compact_dataset_metrics}) validates that this adaptive strategy effectively captures the complex structures and shedding dynamics that challenge fixed kernel architectures.

\subsection{Metrics}

\subsubsection{Velocity Magnitude RMSE}
\label{app:velocity_rmse}
For both Burgers' equation and Navier-Stokes flows, we compute the root mean square error (RMSE) of the predicted velocity magnitude on a domain tessellated into an $H \times W$ grid:
\begin{equation}
    \text{RMSE}_u = \sqrt{ \frac{1}{HW} \sum_{i=1}^{H} \sum_{j=1}^{W} |u_{ij} - \hat{u}_{ij}|^2 }
\end{equation}
Here, $u_{ij}$ and $\hat{u}_{ij}$ are the velocity magnitudes predicted by the model and the corresponding ground truth at grid point $(i,j)$, respectively. The velocity magnitude at each $(i,j)$ grid point is computed from the directional velocity components as:
\begin{equation}
    u_{ij} = \sqrt{ u_x^2 + u_y^2 }
\end{equation}
where $u_x$ and $u_y$ are the velocity components in the $x$ and $y$ directions. For Burgers' equation, only the relevant velocity component(s) are used based on the problem dimensionality.

\subsubsection{Pressure RMSE}
\label{app:pressure_rmse}
For Navier-Stokes flows and energetic materials, the RMSE of the predicted pressure field is computed as:
\begin{equation}
    \text{RMSE}_P = \sqrt{ \frac{1}{HW} \sum_{i=1}^{H} \sum_{j=1}^{W} |P_{ij} - \hat{P}_{ij}|^2 }
\end{equation}
where $P_{ij}$ is the predicted pressure at grid location $(i,j)$, and $\hat{P}_{ij}$ is the ground truth pressure derived from direct numerical simulation.

\subsubsection{Temperature RMSE}
\label{app:temperature_rmse}
For energetic materials, the RMSE of the predicted temperature field is computed as:
\begin{equation}
    \text{RMSE}_T = \sqrt{ \frac{1}{HW} \sum_{i=1}^{H} \sum_{j=1}^{W} |T_{ij} - \hat{T}_{ij}|^2 }
\end{equation}
where $T_{ij}$ is the predicted temperature at grid location $(i,j)$, and $\hat{T}_{ij}$ is the ground truth temperature from direct numerical simulation.

\subsubsection{Hotspot Metrics for Energetic Materials}
\label{app:hotspot_metrics}
For the energetic materials dataset, standard field-level RMSE metrics (Section~\ref{app:temperature_rmse}) provide an incomplete picture of model performance because they do not capture the critical spatial localization of hotspots regions. We therefore employ spatially-aware metrics that explicitly account for both localization accuracy and temperature prediction quality.

We define hotspots as spatial locations where $T > T_{\text{thresh}} = 875$ K, a threshold corresponding to the onset of rapid thermochemical reaction in our energetic material simulations~\cite{nguyen2022multi}. We treat hotspot prediction as a binary segmentation task, where ground truth and predicted temperature fields are thresholded to produce binary masks $M_{\text{gt}}, M_{\text{pred}} \in \{0,1\}^{H \times W}$.

We compute standard segmentation metrics to quantify spatial alignment. The \textbf{Intersection over Union (IoU)}, measures overlap between predicted and ground truth hotspot regions:
\begin{equation}
    \text{IoU} = \frac{|M_{\text{gt}} \cap M_{\text{pred}}|}{|M_{\text{gt}} \cup M_{\text{pred}}|} = \frac{\text{TP}}{\text{TP} + \text{FP} + \text{FN}}
\end{equation}
where TP (true positives) are pixels correctly identified as hotspots, FP (false positives) are pixels incorrectly identified as hotspots, and FN (false negatives) are hotspot pixels missed by the model. IoU ranges from 0 (no overlap) to 1 (perfect overlap)..

The \textbf{Dice coefficient}, equivalent to the F1 score in binary segmentation, emphasizes overlap more strongly than IoU:
\begin{equation}
    \text{Dice} = \frac{2|M_{\text{gt}} \cap M_{\text{pred}}|}{|M_{\text{gt}}| + |M_{\text{pred}}|} = \frac{2 \cdot \text{TP}}{2 \cdot \text{TP} + \text{FP} + \text{FN}}
\end{equation}
Dice ranges from 0 to 1, with higher values indicating better spatial agreement. It is more sensitive to correct predictions in small hotspot regions compared to IoU.

Beyond spatial overlap, we evaluate how accurately models predict temperature values within correctly identified hotspot regions (true positive pixels). This is critical because even spatially accurate hotspot localization may exhibit significant temperature errors that affect downstream safety analyses. We compute the \textbf{Root Mean Square Error in True Positive Regions (RMSE-TP)}:
\begin{equation}
    \text{RMSE}_{\text{TP}} = \sqrt{\frac{1}{|\text{TP}|} \sum_{(i,j) \in \text{TP}} \left(T_{\text{pred}}(i,j) - T_{\text{gt}}(i,j)\right)^2}
\end{equation}
where $\text{TP} = \{(i,j) : M_{\text{gt}}(i,j) = 1 \land M_{\text{pred}}(i,j) = 1\}$ is the set of true positive pixels. By restricting the error metric to true positive regions, we isolate temperature prediction accuracy from spatial localization errors. Lower RMSE-TP indicates more accurate temperature prediction within correctly detected hotspots.

To evaluate joint spatial-temperature performance, we introduce an \textbf{IoU-weighted temperature error} that balances spatial localization and thermal prediction accuracy:
\begin{equation}
    \text{IoU-weighted RMSE} = \text{RMSE}_{\text{TP}} \times (2 - \text{IoU})
\end{equation}
This metric amplifies temperature errors when spatial overlap is poor. When $\text{IoU} = 1$ (perfect spatial alignment), the metric equals RMSE-TP; when $\text{IoU} = 0$ (no spatial overlap), temperature error is doubled. The linear weighting $(2 - \text{IoU})$ ensures that improvements in either spatial overlap or temperature accuracy contribute proportionally to the combined score, without requiring arbitrary normalization constants.

\end{document}